# Recent Advances in Multi-Choice Machine Reading Comprehension: A Survey on Methods and Datasets

Shima Foolad[a], Kourosh Kiani[a,*], Razieh Rastgoo[a]

[a]*Department of Electrical & Computer Engineering, Semnan University, Semnan, Iran*

**Abstract**

This paper provides a thorough examination of recent developments in the field of multi-choice Machine Reading Comprehension (MRC). Focused on benchmark datasets, methodologies, challenges, and future trajectories, our goal is to offer researchers a comprehensive overview of the current landscape in multi-choice MRC. The analysis delves into 30 existing cloze-style and multiple-choice MRC benchmark datasets, employing a refined classification method based on attributes such as corpus style, domain, complexity, context style, question style, and answer style. This classification system enhances our understanding of each dataset's diverse attributes and categorizes them based on their complexity. Furthermore, the paper categorizes recent methodologies into Fine-tuned and Prompt-tuned methods. Fine-tuned methods involve adapting pre-trained language models (PLMs) to a specific task through retraining on domain-specific datasets, while prompt-tuned methods use prompts to guide PLM response generation, presenting potential applications in zero-shot or few-shot learning scenarios. By contributing to ongoing discussions, inspiring future research directions, and fostering innovations, this paper aims to propel multi-choice MRC towards new frontiers of achievement.

## 1. Introduction

Reading comprehension, a fundamental human skill, typically involves analyzing a text and responding to questions about its content. Effectively answering these questions requires extracting relevant information from the text. Transferring this skill to machines is a major challenge in artificial intelligence, as its successful implementation could revolutionize natural language processing (NLP) applications by enabling them to operate on a question-answering paradigm, similar to ChatGPT (Ouyang et al., 2022). The rise of large language models (LLMs) like ChatGPT has revolutionized search, enabling machines to provide human-like answers to complex questions. This shift is evident in Fig. 1, which demonstrates the ability of ChatGPT to answer questions accurately and creatively. Machine reading comprehension (MRC) plays a crucial role in this process, as it allows LLMs to grasp the nuances of the query and formulate a

---

* Corresponding author

*Email addresses:* sh.foolad@semnan.ac.ir (S. Foolad), kourosh.kiani@semnan.ac.ir (K. Kiani)

comprehensive response.

MRC systems have made significant progress in recent years, particularly in the domain of multi-choice datasets. The multi-choice variant of MRC presents a demanding challenge, requiring models to identify the correct answer from a set of options. Multi-choice MRC tasks involve natural-form questions, while cloze-style MRC tasks present incomplete questions with missing sections that must be filled in using the provided options. This comprehensive survey delves into the recent advancements in multi-choice and cloze-style MRC, exploring key methodologies, benchmark datasets, recent models, and the challenges faced by researchers in this field. Additionally, we provide a thorough evaluation of these datasets and existing methods to inform future research directions.

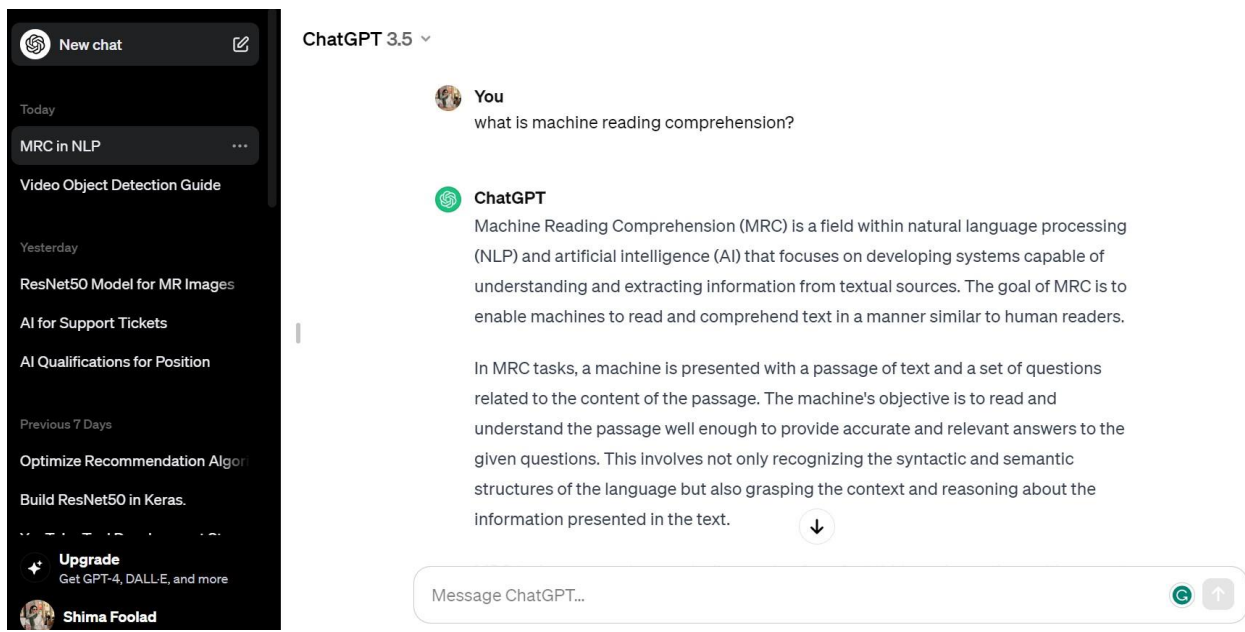

**Fig. 1:** A ChatGPT demo to answer the questions (https://chat.openai.com/)

We believe that our survey will be a valuable resource for researchers and practitioners working on MRC. This survey makes several significant contributions to this field:

- **Benchmark Datasets Analysis:** We provide an in-depth analysis of 30 existing cloze-style and multiple-choice MRC benchmark datasets, some of which have been underexplored in the community. This comprehensive survey covers aspects such as dataset statistics, data sources, availability, evaluation metrics, and human vs. model performance.
- **Refined Classification Method:** We propose a refined classification method for cloze-style and multi-choice MRC datasets based on corpus style, domain, complexity, context style, question style, and answer style. This classification provides a more comprehensive understanding of the diverse attributes of each dataset and categorizes them according to their complexity, a challenging issue in recent datasets.
- **Methodologies Exploration:** We present a comprehensive exploration of the recent methodologies and the state-of-the-art models in MRC, focusing on fine-tuned and prompt-tuned methods. We detail the process of fine-tuning and prompt-tuning, highlighting their advantages and potential for zero-shot or few-shot learning scenarios.

- **Challenges Identification:** Despite the remarkable progress achieved in MRC, we identify several challenges that remain. These challenges, including data scarcity and bias, complex reasoning, commonsense knowledge utilization, explanation and interpretability, domain adaptation and cross-lingual transfer, computational cost and scalability, robustness to noise and errors, and handling unseen answer types and ambiguous questions, hinder further development and applicability of these methods in real-world scenarios.
- **Future Directions Outline:** We outline potential future directions for research in the field of MRC, such as enhancing data representation, improving complex reasoning, incorporating commonsense knowledge, enhancing explainability, optimizing computational efficiency, and better handling of noisy data and ambiguous and unseen answer types.

The rest of this document is structured as follows: Section 2 examines cloze-style and multiple-choice MRC benchmark datasets. Section 3 provides a thorough exploration of recent methodologies in MRC. Sections 4 and 5 respectively outline the main challenges and future directions in the field of MRC. The conclusion is presented in Section 6.

## 2. Benchmark Datasets

The benchmark datasets play a crucial role in speeding up the development of better neural models. In the past few years, the number of MRC datasets has increased exponentially. These novel datasets inspired a huge number of new language models, such as RoBERTa (Y. Liu et al., 2019), ERNIE (S. Wang et al., 2021), T5 (Raffel et al., 2019), DeBERTa (P. He et al., 2020), and PaLM (Chowdhery et al., 2022). The performance of the state-of-the-art (SOTA) models has already exceeded human performance over the related MRC benchmark datasets.

Most researchers focus on a few popular MRC datasets while most other MRC datasets are not widely known and studied by the community. To address these gaps and determine which tasks were solved, we list a comprehensive survey of existing cloze style and multi-choice MRC benchmark datasets in Table 1, including quantitative analysis, data sources, dataset link, evaluation metrics along with Human and SOTA model performances. We will discuss each of these aspects.

**Table 1.** Details of multi-choice MRC datasets.

| year | Dataset | Statistics | | | Data Sources | Evaluation | | | | Dataset Link |
|---|---|---|---|---|---|---|---|---|---|---|
| | | Vocab Size | Question Size | Context Size | | Metrics | Human (%) | SOTA (%) | Solved | |
| 2016 | CBT (Hill et al., 2015) | 53.6k | 687k | 108 books | Stories, Narrative | ACC | 81.6 | 74.9 | ✗ | url |
| 2016 | MovieQA (Tapaswi et al., 2015) | 21k | 15k | 408 movies | Plots, Subtitles, Video clips, Scripts, and DVS | ACC | - | 42.53 | - | url |
| 2016 | WDW (Onishi et al., 2016) | 347k | 330k | 206k | Gigaword | ACC | 84 | 72.6 | ✗ | url |
| 2017 | BookTest (Bajgar et al., 2016) | 200k | 14M | 13.5k books | Project Gutenberg | ACC | 81.6 | 83.7 | ✓ | url |
| 2017 | Quasar-S (Dhingra et al., 2017) | 987k | 37k | - | Stack Overflow | ACC | 50 | 33.6 | ✗ | url |
| 2017 | WikiHop (Welbl et al., 2018) | 304k | 51k | 49k | Wikipedia | ACC | - | 84.4 | - | url |

| Year | Dataset | | | | | | | | | |
|---|---|---|---|---|---|---|---|---|---|---|
| 2017 | RACE (Lai et al., 2017) | 136k | 97k | 27k | English exam | ACC | 94.5 | 91.4 | ✗ | url |
| 2017 | SciQ (Welbl et al., 2017) | 23k | 13k | 13k | Science exam | ACC | 87.8 | 96.6 | ✓ | url |
| 2018 | RecipeQA-text (Yagcioglu et al., 2018) | 63k | 36k | 19k recipe | Instructibles.com | ACC | 73.6 | 47.5 | ✗ | url |
| 2018 | CliCR (Šuster & Daelemans, 2018) | 122k | 105k | 12k | BMJ Clinical Case Reports | F1/EM | 53.7 / 35 | 33.9/24.5 | ✗ | url |
| 2018 | CLOTH (Xie et al., 2018) | 37k | 99k | 7k | English Exam | ACC | 85.9 | 70.7 | ✗ | url |
| 2018 | ReCoRD (S. Zhang et al., 2018) | 139k | 121k | 73k | News article | F1/EM | 91.7 / 91.3 | 96.4/95.9 | ✓ | url |
| 2018 | BioRead (Pappas et al., 2018) | 3.9M | 16.4M | 16.4M | PubMed Central | ACC | - | 51.19 | - | url |
| 2018 | MultiRC (Khashabi et al., 2018) | 23k | 6k | 871 | News and other web pages | F1 | 81.8 | 89.6 | ✓ | url |
| 2018 | ARC (Clark et al., 2018) | 4M | 7k | 14M | Science exam | ACC | - | 86.9 | - | url |
| 2018 | MedQA (Jin et al., 2020) | - | - | 243k | Medical exam | ACC | - | 75.3 | - | - |
| 2018 | MCScript (Ostermann et al., 2018) | 8k | 32k | 2k | Scripts, CRW | ACC | 98.2 | 84.8 | ✗ | - |
| 2019 | MCScript2.0 (Ostermann et al., 2019) | 12k | 20k | 3k | Narrative | ACC | 97.0 | - | ✗ | - |
| 2019 | RACE-C (Liang et al., 2019) | 58k | 14k | 4k | English exam | ACC | - | - | - | url |
| 2019 | DREAM (Sun et al., 2019) | 10k | 10k | 6k | English exam | ACC | 98.6 | 92.6 | ✗ | url |
| 2019 | CosmosQA (Huang et al., 2019) | 40k | 35.6k | 21.9k | Blogs, Narrative | ACC | 94 | 91.7 | ✗ | url |
| 2019 | Shmoop (Chaudhury et al., 2019) | - | 7k | 7k | Project Gutenberg | ACC | - | 40 | - | url |
| 2020 | BioMRC (Stavropoulos et al., 2020) | - | 812k | 812k | PUBTATOR | ACC | - | 88 | - | url |
| 2020 | ReClor (Yu et al., 2020) | 17k | 6k | 6k | Exam | ACC | 63.0 | 90.2 | ✓ | url |
| 2020 | QuAIL (Rogers et al., 2020) | 17k | 15k | 800 | News, Stories, Fiction, Blogs | ACC | 60.0 | 53.4 | ✗ | url |
| 2020 | QASC (Khot et al., 2019) | 1.6M | 10k | 17M | Science exam, WorldTree | ACC | 93 | 93 | ✓ | url |
| 2020 | LogiQA (J. Liu et al., 2020) | - | 8.6k | 8.6k | Chinese Civil Service Exam | ACC | 86/95 | 49.17 | ✗ | url |
| 2022 | ExpMRC-RACE+ (Cui et al., 2021) | - | 1k | 335 | RACE dataset | ACC | 84.4 | 50.86 | ✗ | url |
| 2023 | LogiQA2.0 (H. Liu et al., 2023) | - | 15.7k | 15.7k | Chinese Civil Service Exam | ACC | 84/98 | 54.93 | ✗ | url |
| 2023 | RULE (Kawabata & Sugawara, 2023) | 9K | 4K | 943 | Exam | ACC | 81.5 | 69.2 | ✗ | url |

## 2.1 Dataset Statistics

In Table 1, a comprehensive overview of per-dataset statistics is presented, encompassing vocabulary size, question size, and context size. These metrics serve as valuable indicators for estimating the computational requirements associated with training MRC system. Fig. 2 further illuminates the distinctions between datasets, specifically in terms of question size (dataset size). Among the 30 datasets examined, only five (CBT (Hill et al., 2015), BookTest (Bajgar et al., 2016), BioRead (Pappas et al., 2018), BioMRC (Stavropoulos et al., 2020), WDW (Onishi et al., 2016)) exceed 140k questions, rendering them particularly suitable for training or fine-tuning deep learning models. Notably, CBT and BookTest, although older, still stand out. BioRead and BioMRC operate within the medical domain, while WDW remains inaccessible to the public. Datasets with question sizes ranging from 40k to 140k, such as RACE (Lai et al., 2017), CliCR (Šuster & Daelemans, 2018), CLOTH (Xie et al., 2018), ReCoRD (S. Zhang et al., 2018), and WikiHop (Welbl et al., 2018) are also popular choices. Additionally, several datasets with less than 40k questions may be ideal for transfer learning on emerging language models. In light of the evolving landscape where new language models showcase multitasking capabilities, recent benchmarks like SuperGLUE (A. Wang et al., 2019) have emerged for evaluating general-purpose language understanding models. SuperGLUE, featuring two MRC tasks—ReCoRD and MuliRC (Khashabi et al., 2018) — among its eight tasks, is illustrated in Fig. 3. Notably, multi-task datasets like SuperGLUE have garnered increased attention from researchers compared to single-task datasets like CBT and RACE. This shift underscores the growing importance of evaluating models across diverse linguistic tasks and domains.

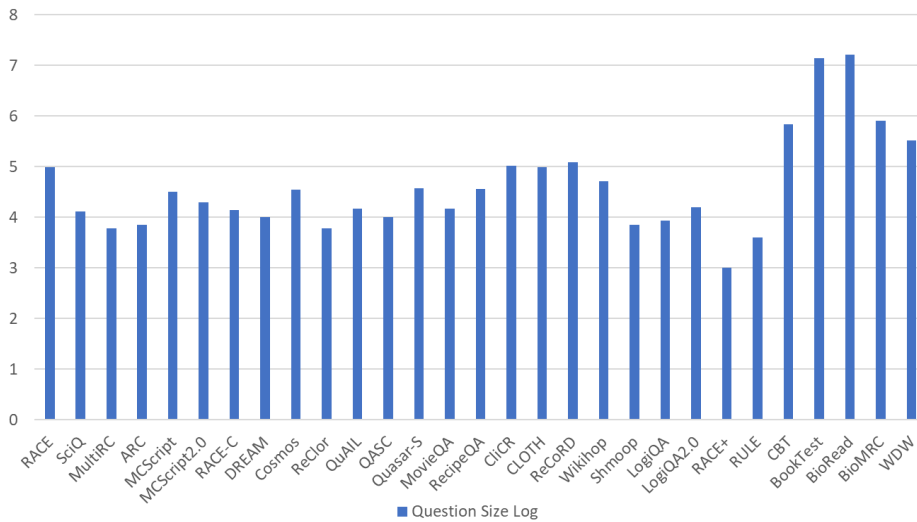

**Fig. 2.** Differences between the datasets in terms of question size. To facilitate better visualization and comparison, the question size is presented in logarithmic scale.

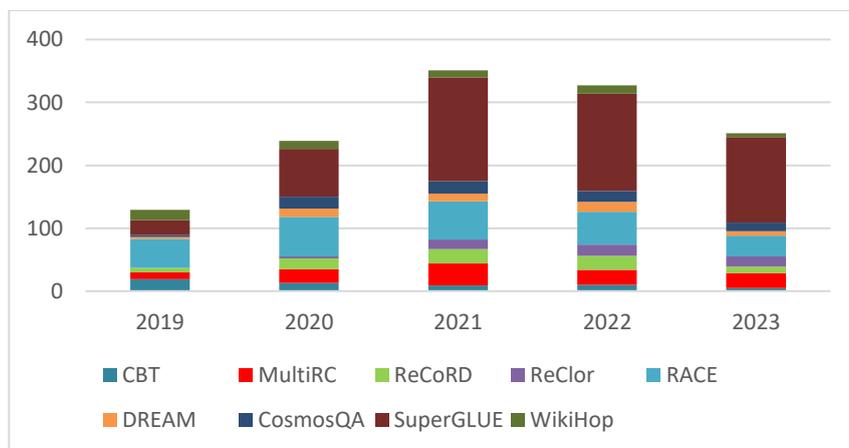

**Fig. 3.** The number of research papers in paperswithcode.com for most used datasets

## 2.2 Data Sources

A substantial number of the datasets outlined in Table 1 (13 out of 30) predominantly utilize exams as their primary data source, while others make use of web pages, news articles, books, and narratives. The distribution of these diverse data sources across datasets is depicted in Fig. 4.

- **Exam:** The RACE, RACE-C (Liang et al., 2019), CLOTH, and DREAM (Sun et al., 2019) datasets were gathered from college English examinations. RACE-C was combined with RACE-M and RACE-H, collected from middle and high school exams, to expand the RACE dataset. Additionally, the RACE$^+$ dataset, a newly annotated subset of the ExpMRC benchmark (Cui et al., 2021), includes extra hints about the answering process, offering valuable insights for evidence annotation. ReClor (Yu et al., 2020) and RULE (Kawabata & Sugawara, 2023) utilizes exams from GMAT and LSAT, LogiQA (J. Liu et al., 2020) and LogiQA2.0 (H. Liu et al., 2023) from the Chinese Civil Service Examination, MedQA (Jin et al., 2020) from medical exams, and ARC (Clark et al., 2018), QASC (Khot et al., 2019), and SciQ (Welbl et al., 2017) from science exams.
- **Web Pages:** The Quasar-S (Dhingra et al., 2017) dataset was collected by extracting software entity tags from the Stack Overflow website. WikiHop is sourced from Wikipedia, RecipeQA (Yagcioglu et al., 2018) is derived from Instructables[1], and QuAIL (Rogers et al., 2020) is compiled from various blogs, including content from platforms like Quora, Voanews, and Manybooks web pages[2]. The CliCR, BioRead, and BioMRC datasets belong to the biomedical domain and were sourced from BMJ Case Reports[3], PubMed Central[4] and PUBTATOR, respectively.
- **News:** The MultiRC and ReCoRD datasets were collected from news sources, while WDW utilizes Gigaword as its data source. QuAIL includes political news from Voice of America.
- **Book:** The books consist of Project Gutenberg for CBT, BookTest and shmoop (Chaudhury et al., 2019) datasets, as well as fiction published under CC license for QuAIL.
- **Stories:** The CBT, MCScript (Ostermann et al., 2018), MCScript2.0 (Ostermann et al., 2019) and CosmosQA (Huang et al., 2019) were gathered from narrative stories, while MovieQA (Tapaswi et al., 2015) was sourced from movie narratives. QuAIL, on the other hand, consists of user stories published on Quora.

## 2.3 Availability

Within the datasets detailed in Table 1, only 41.37% include leaderboards and corresponding dataset links for the evaluation and comparison of models. Table 2 provides the download and leaderboard links for these datasets. As you can see in Table 2, the ReCoRD and WikiHop datasets refrain from releasing test data publicly to ensure the preservation of test result integrity.

---

[1] https://www.instructables.com/
[2] https://www.quora.com/about/to, https://www.voanews.com/, http://manybooks.net/categories/CCL
[3] http://casereports.bmj.com/
[4] https://www.ncbi.nlm.nih.gov/pmc/tools/openftlist/

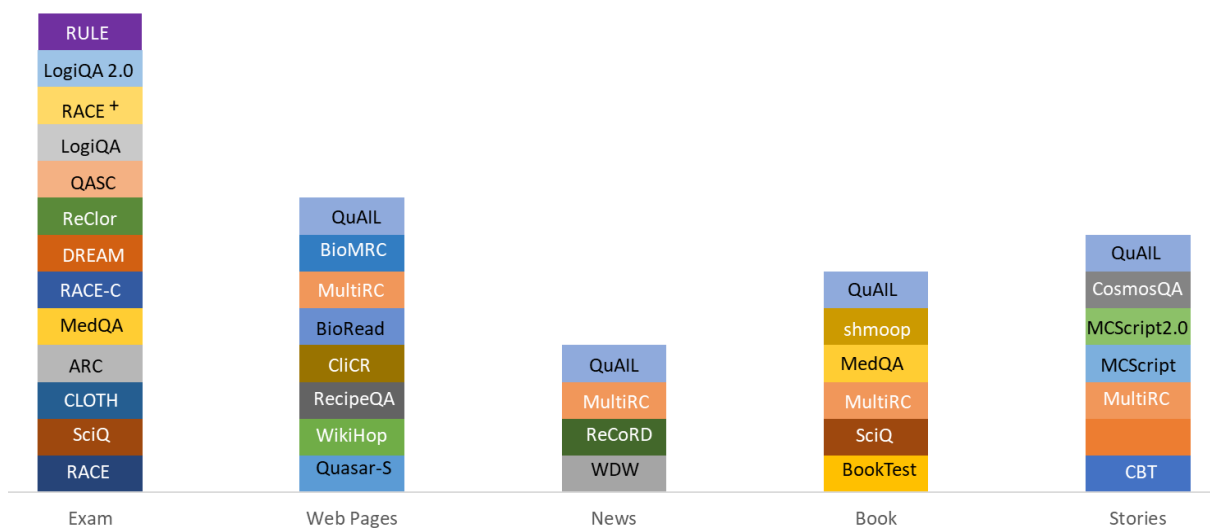

**Fig. 4.** Distribution of data sources across datasets

**Table 2:** The available download and leaderboard links of the datasets detailed in Table 1.

| Year | Dataset | Download link | Leaderboard link | Test data availability |
|---|---|---|---|---|
| 2017 | RACE (Lai et al., 2017) | http://www.cs.cmu.edu/~glai1/data/race/RACE.tar.gz | http://www.qizhexie.com/data/RACE_leaderboard.html | ✓ |
| 2017 | WikiHop (Welbl et al., 2018) | http://bit.ly/2m0W32k | https://qangaroo.cs.ucl.ac.uk/leaderboard.html | ✗ |
| 2018 | MultiRC (Khashabi et al., 2018) | https://dl.fbaipublicfiles.com/glue/superglue/data/v2/MultiRC.zip | https://super.gluebenchmark.com/leaderboard | ✓ |
| 2018 | ARC (Clark et al., 2018) | https://ai2-public-datasets.s3.amazonaws.com/arc/ARC-V1-Feb2018.zip | https://leaderboard.allenai.org/arc/submissions/public | ✓ |
| 2018 | RecipeQA-text (Yagcioglu et al., 2018) | https://hucvl.github.io/recipeqa/ | https://hucvl.github.io/recipeqa/ | ✓ |
| 2018 | ReCoRD (S. Zhang et al., 2018) | https://sheng-z.github.io/ReCoRD-explorer/ | https://sheng-z.github.io/ReCoRD-explorer/ https://super.gluebenchmark.com/leaderboard | ✗ |
| 2019 | DREAM (Sun et al., 2019) | https://github.com/nlpdata/dream | https://dataset.org/dream | ✓ |
| 2019 | CosmosQA (Huang et al., 2019) | https://github.com/wilburOne/cosmosqa/tree/master/data/ | https://leaderboard.allenai.org/cosmosqa/submissions/public | ✓ |
| 2020 | ReClor (Yu et al., 2020) | https://github.com/yuweihao/reclor/releases/download/v1/reclor_data.zip | https://eval.ai/web/challenges/challenge-page/503/leaderboard/1347 | ✓ |
| 2020 | QuAIL (Rogers et al., 2020) | https://github.com/text-machine-lab/quail/ | https://text-machine.cs.uml.edu/lab2/projects/quail/ | ✓ |
| 2020 | QASC (Khot et al., 2019) | https://ai2-public-datasets.s3.amazonaws.com/qasc/qasc_dataset.tar.gz | https://leaderboard.allenai.org/qasc/submissions/public | ✓ |
| 2022 | ExpMRC-RACE+ (Cui et al., 2021) | https://github.com/ymcui/expmrc/tree/main/data/race | https://ymcui.com/expmrc/ | ✓ |

## 2.4 Evaluation

The evaluation of multi-choice and cloze datasets typically relies on accuracy. Accuracy (ACC) measures the proportion of questions for which the model correctly identifies the answer. However, for datasets like CliCR and ReCoRD, which require the model to generate or extract entity-based answers, Exact Match (EM) and F1 are more appropriate metrics. In these datasets, the correct answer is selected from multiple candidate options, and EM is equivalent to ACC. EM is considered a stricter metric than ACC and is often used in conjunction with other metrics, such as F1 (which considers partial correctness), to provide a more comprehensive evaluation of MRC systems. F1 measures the overlap between the model's generated answer and the correct answer in terms of word tokens. For datasets like MultiRC, which have multiple correct answers, F1 is the preferred metric.

Some datasets provide valuable insights through reported human performance, as shown in Table 1. This human performance perspective provides a nuanced understanding of the complexity of the questions within the dataset. The contrast between the comparatively lower human performance scores in Quasar-S, CliCR, and QuAIL and the notably higher scores in DREAM and MC-Script is striking. Human performance serves as a critical reference point for automated systems, enabling researchers to identify less-explored datasets where there are significant gaps between cutting-edge machine performance and human capabilities, thereby encouraging further investigation. To facilitate easier comparison and identification of datasets with substantial gaps between human performance and the State-Of-The-Art (SOTA) model, we have provided the performance of SOTA models for each dataset in Table 1. For datasets in Table 1 with available leaderboard links, we obtained SOTA performance from the leaderboard, and for others, we referred to paperswithcode.com. In cases where the SOTA model for a dataset was not explicitly mentioned in search results, we conducted a thorough search on online platforms like arXiv.org and ACL Anthology. To identify datasets that still pose significant challenges and have not yet achieved human-level performance, we introduced a "solved" column in Table 1, marking which datasets remain unsolved. The symbol "χ" indicates that the dataset has not been fully solved, while "✓" indicates that there exists a model that has achieved higher performance than humans on the dataset. For a more intuitive comparison between SOTA and human performance for each dataset, refer to the visualization in Fig. 5.

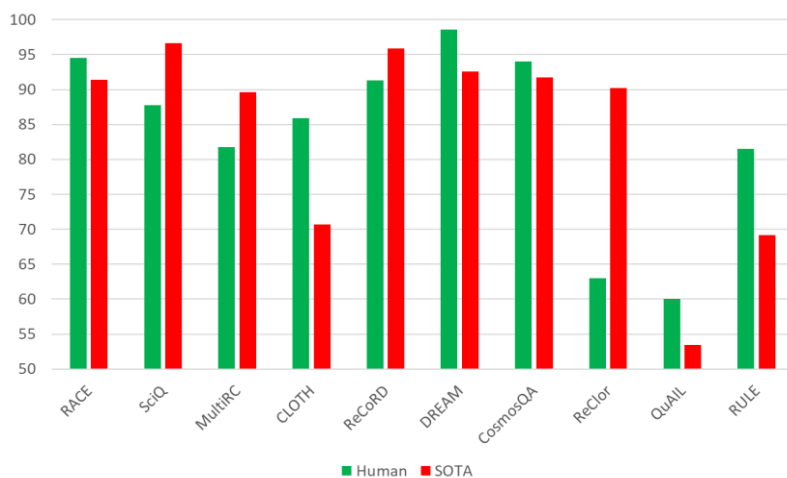

**Fig. 5.** Comparing SOTA and Human performance of MRC datasets

## 2.5 Datasets Classification

Several researchers (S. Liu et al., 2019; Qiu et al., 2019) traditionally categorize MRC datasets into four groups: cloze style, multiple-choice, span prediction, and free form. However, this classification method lacks precision, as certain MRC datasets can simultaneously fall into both cloze style and multiple-choice style categories. For instance, CBT, BookTest, and CLOTH (as depicted in Fig. 6) exemplify this dual classification. Take CBT as an example: although its questions are cloze-style, the system is presented with 10 entity candidates for selecting the correct answer, making it a multi-choice dataset as well. The distinguishing factor for these cloze multi-choice datasets compared to other multi-choice datasets lies in the question form—in cloze-style datasets, the answer represents a missing part of the question, whereas in other multi-choice datasets, the questions have no missing words.

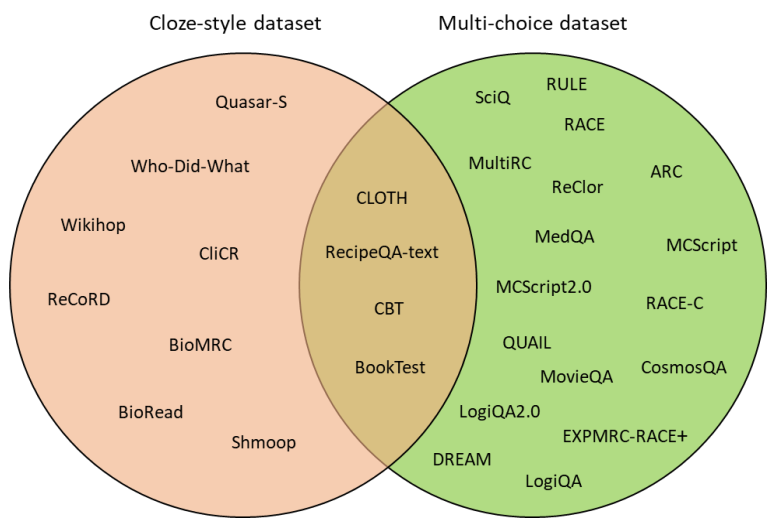

**Fig. 6.** Classifying MRC datasets to cloze-style and multi-choice categories

Recently, some researchers (Baradaran et al., 2020; Dzendzik et al., 2021; Zeng et al., 2020) have proposed a categorization based on the corpus type, question type, answer type, and context type of the datasets, which appears to be a more acceptable approach. Recognizing that the variation among datasets lies in different forms of questions, answers, and their corpora, we delved deeper into dataset details. In addition to considering question, answer, and corpus types, we incorporated other crucial factors such as dataset complexity. Fig. 7 illustrates our refined classification method for cloze-style and multi-choice MRC datasets. We organized the datasets into six distinct categories to provide a comprehensive understanding of their diverse attributes: corpus style, domain, complexity, context style, question style, and answer style. Subsequently, we will delve into a detailed explanation of each group.

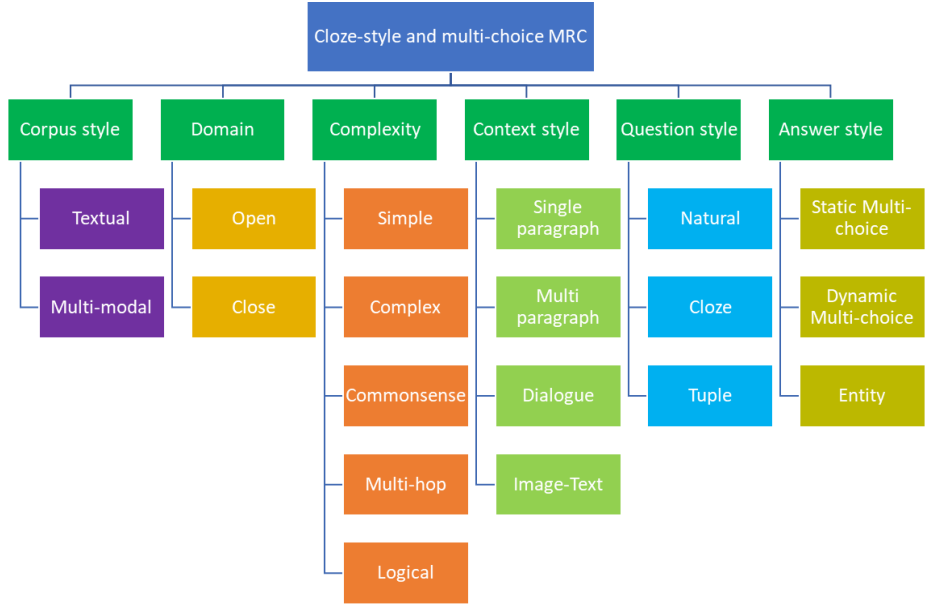

**Fig. 7.** A new classification method of cloze-style and multi-choice MRC datasets

In terms of corpus style, MRC datasets are divided into textual and multi-modal categories, with most datasets falling into the former while emerging datasets like MovieQA and RecipeQA introduce a multimodal dimension by incorporating images and videos alongside text.

Concerning domain, MRC datasets are classified as open-domain or closed-domain. Notably, open domain datasets such as WikiHop cover diverse subjects, while closed domain datasets, exemplified by CliCR, BioRead, and BioMRC, focus on specific domains like biomedicine.

Regarding complexity, MRC datasets are divided into five categories, each representing different levels of difficulty for models. The first category, "Simple" pertains to datasets where questions demand straightforward comprehension of provided context. If the questions primarily involve direct retrieval of information from the given context without intricate reasoning or inference, it would be reasonable to include them in the "Simple" category such as CBT and BookTest. Conversely, if they involve more complex reasoning or understanding of implicit information, they belong to "Complex" category. The "Complex" category exemplified by datasets like RACE and CLOTH, involves more intricate analysis and integration of information. "Commonsense" the third category, is characterized by datasets like ReCoRD and DREAM, where questions extend beyond explicit text and necessitate a grasp of implicit, real-world knowledge. The fourth category, "Multi-hop" features datasets like wikiHop and MultiRC are challenging tasks where the model needs to "hop" between multiple pieces of information to navigate and synthesize information across multiple passages for accurate responses. Finally, the fifth category, "Logical" is illustrated by datasets like ReClor and LogiQA, emphasizing the need for systematic application of formal logic in deriving answers. This categorization provides a nuanced framework for evaluating and understanding the diverse cognitive skills demanded by MRC datasets, contributing to a more comprehensive assessment of the capabilities of machine comprehension models across varying degrees of complexity. We categorized the datasets detailed in Table 1 according to their complexity, visualizing the results in a sunburst chart depicted in Fig. 8. In this graphical representation, the orange segment signifies datasets characterized by simple reasoning, the blue segment denotes those involving complex reasoning, the green segment represents datasets requiring commonsense reasoning, the yellow segment is allocated to datasets emphasizing logical reasoning, and the dark red section corresponds to datasets highlighting multi-hop reasoning. A notable observation is the substantial prevalence of datasets demanding complex reasoning, with comparatively smaller proportions assigned to logical reasoning and multi-hop reasoning.

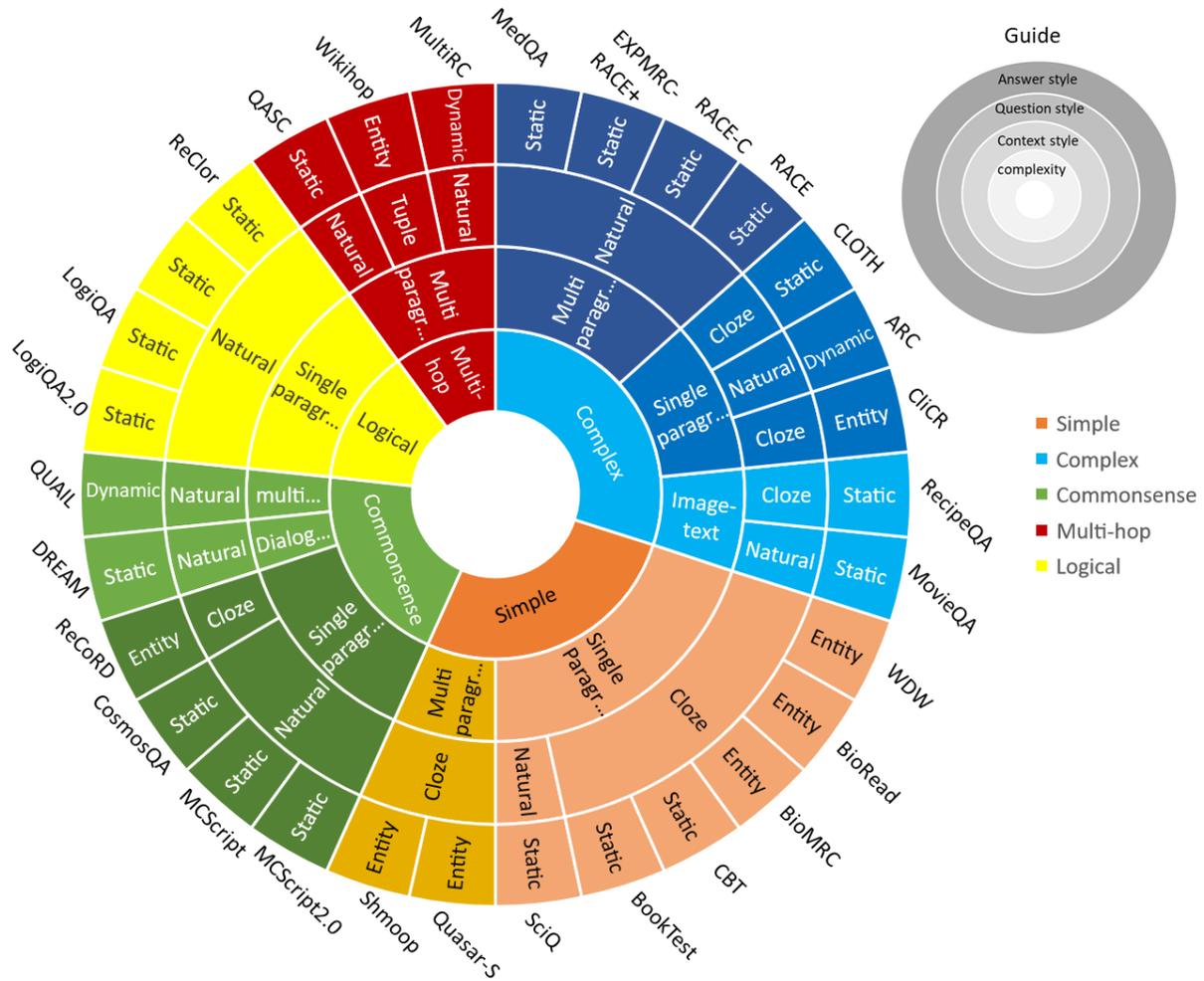

**Fig. 8.** A chart representing complexity type of the MRC datasets along with other styles of new categories.

Beyond complexity, our visualization in Fig. 8 also provides insight into the context, question, and answer styles of each dataset, presented in distinct layers of rings. The context style varies, ranging from dialogues as seen in DREAM to multimodal setups like the image-text combination in MovieQA and RecipeQA datasets or the more conventional natural text. We further categorized datasets with natural contexts into single or multi-paragraph structures, typically observed in datasets emphasizing multi-hop reasoning.

Regarding question types, we classify them into three categories: Cloze, Tuple, and Natural. In Cloze-type questions, exemplified by ReCoRD, the question is structured as a sentence with a missing word or phrase, requiring insertion or completion. Tuple-type questions, as in WikiHop, adopt the format of (?, property, subject) triples, adding a unique dimension to the questioning structure. Natural-type questions retain the inherent structure of a typical query.

The answer style across datasets predominantly follows a multi-choice format, where several candidate options are presented, and the correct answer must be selected. Further categorization of answer styles includes Static, Dynamic, and Entity. In datasets labeled as Static, the number of options is fixed, as indicated in Table 3. For instance, RACE features four candidate answers per question, while CBT has ten. In contrast, datasets categorized as Dynamic witness varying numbers of candidate answers per question, as exemplified by MultiRC, where the count may fluctuate between different instances. Lastly, datasets such as Quasar-S, where the answer is chosen from a set of entity candidates, fall under the Entity group. This detailed categorization enhances our understanding of the nuances in question and answer structures within MRC datasets, contributing to a comprehensive evaluation of machine comprehension models across diverse styles and complexities. For better understanding, Table 4 provides illustrative examples of datasets for each category along with their types.

**Table 3.** defining style of new categories for each dataset in Table 1.

| Dataset | Corpus style | Domain | Complexity (reasoning) | Context style | Question style | Answer style (Multi-choice) | Answer type |
|---|---|---|---|---|---|---|---|
| **CBT** (Hill et al., 2015) | Textual | Open | Complex | Single paragraph | Cloze | Static (10 options) | verb, pronoun, named entity, common noun |
| **RACE** (Lai et al., 2017) | Textual | Open | Complex | Multi paragraph | Natural | Static (4 options) | Phrase, sentence |
| **SciQ** (Welbl et al., 2017) | Textual | Close (science) | Simple | Single paragraph | Natural | Static (4 options) | Word, phrase |
| **MultiRC** (Khashabi et al., 2018) | Textual | Open | Multi-hop | Multi paragraph | Natural | Dynamic | Phrase, sentence |
| **ARC** (Clark et al., 2018) | Textual | Open | Complex | Single paragraph | Natural | Dynamic | Word, phrase, sentence |
| **MedQA** (Jin et al., 2020) | Textual | Close (medical) | Complex | Multi paragraph | Natural | Static (5 options) | Word, phrase |
| **MCScript** (Ostermann et al., 2018) | Textual | Open | Commonsense | Single paragraph | Natural | Static (2 options) | Yes/No, phrase, sentence |
| **MCScript2.0** (Ostermann et al., 2019) | Textual | Close (narrative) | Commonsense | Single paragraph | Natural | Static (2 options) | Yes/No, phrase, sentence |
| **RACE-C** (Liang et al., 2019) | Textual | Open | Complex | Multi paragraph | Natural | Static (4 options) | sentence |
| **DREAM** (Sun et al., 2019) | Textual | Open | Commonsense | Dialogue | Natural | Static (3 options) | Word, phrase, sentence |
| **CosmosQA** (Huang et al., 2019) | Textual | Open | Commonsense | Single paragraph | Natural | Static (4 options) | sentence |
| **ReClor** (Yu et al., 2020) | Textual | Open | Logical | Single paragraph | Natural | Static (4 options) | sentence |
| **QuAIL** (Rogers et al., 2020) | Textual | Open | Commonsense | multi paragraph | Natural | Dynamic | Phrase, sentence |
| **QASC** (Khot et al., 2019) | Textual | Open | Multi-hop | Multi paragraph | Natural | Static (4 options) | Word, phrase |
| **WDW** (Onishi et al., 2016) | Textual | Open | Simple | Single paragraph | Cloze | Entity | person named entity |
| **BookTest** (Bajgar et al., 2016) | Textual | Open | Simple | Single paragraph | Cloze | Static (10 options) | Named entity, common noun |
| **Quasar-S** (Dhingra et al., 2017) | Textual | Open | Simple | Multi paragraph | Cloze | Entity | Named entity |
| **MovieQA** (Tapaswi et al., 2015) | Multi-Modal | Open | Complex | Image, Dialogue | Natural | Static (5 options) | Phrase, sentence |
| **RecipeQA-text** (Yagcioglu et al., 2018) | Multi-Modal | Close (cooking recipe) | Complex | Image, Recipe | Cloze | Static (4 options) | phrase |
| **CliCR** (Šuster & Daelemans, 2018) | Textual | Close (clinical) | Complex | Single paragraph | Cloze | Entity | Clinical entity |
| **CLOTH** (Xie et al., 2018) | Textual | Open | Complex | Single paragraph | Cloze | Static (4 options) | word |
| **ReCoRD** (S. Zhang et al., 2018) | Textual | Open | Commonsense | Single paragraph | Cloze | Entity | Named entity |
| **Wikihop** (Welbl et al., 2018) | Textual | Open | Multi-hop | Multi paragraph | Tuple | Entity | Named entity |
| **BioMRC** (Stavropoulos et al., 2020) | Textual | Close (biomedical) | Simple | Single paragraph | Cloze | Entity | Biomedical entity |
| **BioRead** (Pappas et al., 2018) | Textual | Close (medical) | Simple | Single paragraph | Cloze | Entity | Biomedical entity |
| **Shmoop** (Chaudhury et al., 2019) | Textual | Open | Simple | Multi paragraph | Cloze | Entity | Named entity |
| **LogiQA** (J. Liu et al., 2020) | Textual | Open | Logical | Single paragraph | Natural | Static (4 options) | Sentence |
| **LogiQA2.0** (H. Liu et al., 2023) | Textual | Open | Logical | Single paragraph | Natural | Static (4 options) | Sentence |
| **ExpMRC-RACE**[+] (Cui et al., 2021) | Textual | Open | Complex | Multi paragraph | Natural | Static (4 options) | Phrase, sentence |
| **RULE** (Kawabata & Sugawara, 2023) | Textual | Open | Logical | Single paragraph | Natural | Static (4 options) | sentence |

**Table 4.** A few examples of the MRC datasets based on new categories

| Corpus style | Dataset | Example | |
|---|---|---|---|
| Textual | QASC (Khot et al., 2019) | Context: | Antigens are found on cancer cells and the cells of transplanted organs. Anything that can trigger an immune response is called an antigen. |
| | | Question: | What can trigger immune response? |
| | | Answer: | A. Transplanted organs   B. Desire   C. Pain   D. Death |
| Multi-modal | RecipeQA-text (Yagcioglu et al., 2018) | Context: | Last-Minute Lasagna<br>1. Heat oven to 375 degrees F. Spoon a thin layer of sauce over the bottom of a 9-by-13-inch baking dish.<br>2. Cover with a single layer of ravioli.<br>3. Top with half the spinach half the mozzarella and a third of the remaining sauce.<br>4. Repeat with another layer of ravioli and the remaining spinach mozzarella and half the remaining sauce.<br>5. Top with another layer of ravioli and the remaining sauce not all the ravioli may be needed. Sprinkle with the Parmesan.<br>6. Cover with foil and bake for 30 minutes. Uncover and bake until bubbly, 5 to 10 minutes.<br>7. Let cool 5 minutes before spooning onto individual plates. |
| | | Question: | Choose the best text for the missing blank to correctly complete the recipe Cover. ———. Bake. Cool, serve. |
| | | Answer: | A. Top, sprinkle     B. Finishing touches     C. Layer it up     D. Ravioli bonus round |
| **Domain** | | | |
| Open | Shmoop (Chaudhury et al., 2019) | Context: | **Summary**: Scrooge snorts himself awake, and again it's about to be one o'clock. Scrooge is hip to all this now, though, so he doesn't freak out.<br>**Original Text**: AWAKING in the middle of a prodigiously tough snore, and sitting up in bed to get his thoughts together, Scrooge had no occasion to be told that the bell was again upon the stroke of One. He felt that he was restored to consciousness in the right nick of time, for the especial purpose of holding a conference with the second … |
| | | Question: | Scrooge throws out his famous …. |
| | | Answer: | A. … come over for Christmas dinner, but Scrooge isn't having any of it.<br>B. … but what about the whole Jesus's birth thing?<br>C. … catchphrase - Bah! Humbug!.<br>D. … guys show up asking for any donations for the poor.<br>… |
| Closed | BioMRC (Stavropoulos et al., 2020) | Context: | BACKGROUND: Most brain metastases arise from @entity0 . Few studies compare the brain regions they involve, their numbers and intrinsic attributes. METHODS: Records of all @entity1 referred to Radiation Oncology for … RESULTS: Data from 68 breast and 62 @entity2 @entity1 were compared. Brain metastases presented earlier in the course of the lung than of the @entity0 @entity1 (p = 0.001). There were more metastases in the cerebral hemispheres of the breast than of the @entity2 @entity1 (p = 0.014). More @entity0 @entity1 had cerebellar metastases (p = 0.001). The number of cerebral hemisphere metastases and presence of cerebellar metastases were positively correlated (p = 0.001). The prevalence of at least one @entity3 surrounded with > 2 cm of @entity4 was greater for the lung than for the breast @entity1 (p = 0.019). The @entity5 type, rather than the scanning method, correlated with differences between these variables. CONCLUSIONS: Brain metastases from lung occur earlier, are more @entity4 , but fewer in number than those from @entity0 . Cerebellar brain metastases are more frequent in @entity0 . |
| | | Question: | Attributes of brain metastases from …. |
| | | Candidates: | @entity0 : ['breast and lung cancer'] ; @entity1 : ['patients'] ; @entity2 : ['lung cancer'] ; @entity3 : ['metastasis'] ; @entity4 : ['edematous', 'edema'] ; @entity5 : ['primary tumor'] |
| | | Answer: | @entity0 : ['breast and lung cancer'] |
| **Complexity** | | | |
| Simple | Quasar-S (Dhingra et al., 2017) | Context: | JavaScript is not weakly typed, it is strong typed. JavaScript is a client side Scripting Language. JavaScript was the original client-side web scripting language. |
| | | Question: | javascript – javascript not to be confused with java is a dynamic weakly-typed language used for … as well as server-side scripting . |
| | | Answer: | Client-side |
| Complex | CliCR (Šuster & Daelemans, 2018) | Context: | A gradual improvement in clinical and laboratory status was achieved within 20 days of antituberculous treatment. The patient was then subjected to a thoracic CT scan that also showed significant radiological improvement. Thereafter, tapering of corticosteroids was initiated with no clinical relapse. The patient was discharged after being treated for a total of 30 days and continued receiving antituberculous therapy with no reported problems for a total of 6 months under the supervision of his hometown physicians. |
| | | Question: | If steroids are used, great caution should be exercised on their gradual tapering to avoid … |
| | | Answer: | relapse |
| Commonsense | CosmosQA (Huang et al., 2019) | Context: | It's a very humbling experience when you need someone to dress you every morning, tie your shoes, and put your hair up. Every menial task takes an unprecedented amount of effort. It made me appreciate Dan even more. But anyway, I shan't dwell on this (I'm not dying after all) and not let it detect from my lovely 5 days with my friends visiting from Jersey. |
| | | Question: | What's a possible reason the writer needed someone to dress him every morning? |
| | | Answer: | A. The writer doesn't like putting effort into these tasks. B. The writer has a physical disability. C. The writer is bad at doing his own hair. D. None of the above choices. |
| Multi-hop | Wikihop (Welbl et al., 2018) | Context: | The Troll by Julia Donaldson and David Roberts is a children 's story about a troll and some pirates . The troll in this story is based on the troll from the Three Billy Goats Gruff fairy tale. Meanwhile , the pirate captain Hank Chief and his crew ( Peg Polkadot , Ben Buckle and Percy Patch ) are searching …", "Children's Laureate is a position awarded in the United Kingdom once every two years to a … Ted Hughes and children's writer Michael Morpurgo.", 6 "Julia Catherine Donaldson MBE (born 16 September 1948) is an English writer, playwright and performer, and the 20112013 Children's Laureate. She is best known for her popular rhyming stories for children, especially those illustrated by Axel Scheffler." |
|  |  |  | 7 "Old Norse was a North Germanic language that was spoken by inhabitants of Scandinavia and inhabitants of their overseas settlements during about the 9th to 13th centuries.", |
|  |  | Question: | (country_of_origin, the troll, ?) |
|  |  | Candidates: | africa, denmark, europe, iceland, london, norway, scandinavia, united kingdom |
|  |  | Answer: | <span style="color:red">united kingdom</span> |
| Logical | ReClor (Yu et al., 2020) | Context: | In jurisdictions where use of headlights is optional when visibility is good, drivers who use headlights at all times are less likely to be involved in a collision than are drivers who use headlights only when visibility is poor. Yet Highway Safety Department records show that making use of headlights mandatory at all times does nothing to reduce the overall number of collisions. |
|  |  | Question: | Which one of the following, if true, most helps to resolve the apparent discrepancy in the information above? |
|  |  | Answer: | A. In jurisdictions where use of headlights is optional when visibility is good, one driver in four uses headlights for daytime driving in good weather.<br><span style="color:red">B. Only very careful drivers use headlights when their use is not legally required.</span><br>C. The jurisdictions where use of headlights is mandatory at all times are those where daytime visibility is frequently poor.<br>D. A law making use of headlights mandatory at all times is not especially difficult to enforce. |
| <span style="background-color:#c5e0b4">Context type</span> |  |  |  |
| Single-paragraph | MCScript2.0 (Ostermann et al., 2019) | Context: | We put our ingredients together to make sure they were at the right temperature, preheated the oven, and pulled out the proper utensils. We then prepared the batter using eggs and some other materials we purchased and then poured them into a pan. After baking the cake in the oven for the time the recipe told us to, we then double checked to make sure it was done by pushing a knife into the center. We saw some crumbs sticking to the knife when we pulled it out so we knew it was ready to eat! |
|  |  | Question: | What did they put in the oven? |
|  |  | Answer: | <span style="color:red">A. The cake mix.</span>　　　　B. Utensils. |
| Multi-paragraph | MedQA (Jin et al., 2020) | Context: | Complications: After lithotripsy the majority of patients has transient gross hematuria, which generally does not need special treatment. Renal hematoma formation is relatively rare and can be treated without surgery.<br>One of the characteristics of hematuria caused by kidney tuberculosis is that it frequently happens after a period of bladder irritation, and is more common to have terminal hematuria, which is different from hematuria caused by other urinary diseases.<br>Renal postoperative oliguria or no urine: they are all urological diseases, and can be diagnosed by medical history, physical examination, urology and imaging examinations. It can be clarified whether bilateral hydronephrosis is caused by bladder urinary retention or bilateral ureteral obstruction.<br>… |
|  |  | Question: | Male, 40 years old, gross hematuria with renal colic, ultrasound found stone in right kidney with size 0.6 cm and smooth surface, mild hydronephrosis. The treatment should be: |
|  |  | Answer: | Non-surgical treatment |
| Dialogue | DREAM (Sun et al., 2019) | Context: | W: Tom, look at your shoes. How dirty they are! You must clean them.<br>M: Oh, mum, I just cleaned them yesterday.<br>W: They are dirty now. You must clean them again.<br>M: I do not want to clean them today. Even if I clean them today, they will get dirty again tomorrow.<br>W: All right, then.<br>M: Mum, give me something to eat, please.<br>W: You had your breakfast in the morning, Tom, and you had lunch at school.<br>M: I am hungry again.<br>W: Oh, hungry? But if I give you something to eat today, you will be hungry again tomorrow. |

|  |  | Question: | Why did the woman say that she wouldn't give him anything to eat? |
|---|---|---|---|
|  |  | Answer: | A. Because his mother wants to correct his bad habit.<br>B. Because he had lunch at school.<br>C. Because his mother wants to leave him hungry. |
| Image-Text | MovieQA (Tapaswi et al., 2015) | Context: | Plot: ... secretly betrayed Morpheus to Agent Smith in exchange for a comfortable ...<br>subtitle: 01:04:08 --> 01:04:09<br> ... you know what I realize?<br>01:04:17 --> 01:04:18<br>Ignorance is bliss. 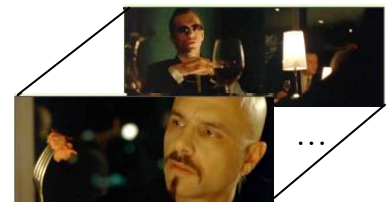 |
|  |  | Question: | Why does Cypher betray Morpheus? |
|  |  | Answer: | In exchange for a comfortable life |
| **Question type** | | | |
| Cloze | ReCoRD (S. Zhang et al., 2018) | Context: | Puerto Rico on Sunday overwhelmingly voted for statehood. But Congress, the only body that can approve new states, will ultimately decide whether the status of the US commonwealth changes. Ninety-seven percent of the votes in …, official results from the State Electoral Commission show. Today, we the people of Puerto Rico are sending a strong and clear message to the US Congress ... and to the world ... claiming our equal rights as American citizens, Puerto Rico Gov. Ricardo Rossello said in a news release and Puerto Rico voted Sunday in favor of statehood. |
|  |  | Question: | For one, they can truthfully say, "Don't blame me, I didn't vote for them," when discussing the … presidency. |
|  |  | Answer: | US |
| Tuple | WikiHop (Welbl et al., 2018) | Context: | The Hanging Gardens, in Mumbai, also known as Pherozeshah Mehta Gardens, are terraced gardens … They provide sunset views over the Arabian Sea …<br>Mumbai is the capital city of the Indian state of Maharashtra. It is the most populous city in India …<br>The Arabian Sea is a region of the northern Indian Ocean bounded on the north by Pakistan and Iran, on the west by northeastern Somalia and the Arabian Peninsula, and on the east by India … |
|  |  | Question: | (Hanging gardens of Mumbai, country, ?) |
|  |  | Answer: | Iran, India, Pakistan, Somalia, … |
| Natural | QuAIL (Rogers et al., 2020) | Context: | The air exploded in a flash of bone and steel and blood. The clash of metal rang through the forest. An arrow pierced through the darkness, its barbed head tearing through flesh and muscle. A roar echoed off of the mountains far to the west. A cry broke through soon after. Then silence … . |
|  |  | Question: | When did the roar happen? |
|  |  | Answer: | A. After the cry<br>B. before the silence<br>C. not enough information to answer this question<br>D. when Char was speaking |
| **Answer type** | | | |
| Entity | WDW (Onishi et al., 2016) | Context: | Tottenham won 2-0 at Hapoel Tel Aviv in UEFA Cup action on Thursday night in a defensive display which impressed Spurs skipper Robbie Keane. ... Keane scored the first goal at the Bloomfield Stadium with Dimitar Berbatov, who insisted earlier on Thursday he was happy at the London club, heading a second. The 26-year-old Berbatov admitted the reports linking him with a move had affected his performances ... Spurs manager Juande Ramos has won the UEFA Cup in the last two seasons ... |
|  |  | Question: | Tottenham manager Juande Ramos has hinted he will allow … to leave if the Bulgaria striker makes it clear he is unhappy. |
|  |  | Candidates: | A. Robbie Keane    B. Dimitar Berbatov |
|  |  | Answer: | B |
| Static multi-choice | RACE (Lai et al., 2017) | Context: | In a small village in England about 150 years ago, a mail coach was standing on the street. It didn't come to that village often …………. .<br>The gentleman was Sir Rowland Hill. He didn't forgot Alice and her letter. "The postage to be paid by the receiver has to be changed," he said to himself and had a good plan. |
| | | | "The postage has to be much lower, what about a penny? And the person who sends the letter pays the postage. He has to buy a stamp and put it on the envelope." he said . The government accepted his plan. Then the first stamp was put out in 1840. It was called the "Penny Black". It had a picture of the Queen on it |
| | | Question 1: | The idea of using stamps was thought of by … |
| | | Candidates: | A. the government<br>B. Sir Rowland Hill<br>C. Alice Brown<br>D. Tom |
| | | Answer: | B |
| | | Question 2: | The first postage stamp was made … |
| | | Candidates: | A. in England<br>B. in America<br>C. by Alice<br>D. in 1910 |
| | | Answer: | A |
| Dynamic multi-choice | MultiRC (Khashabi et al., 2018) | Context: | Sent 1: A group of researchers at a remote jungle island outpost discover the natives are practicing voodoo and black magic.<br>Sent 2: After killing the local priest , a voodoo curse begins to raise the dead to feed on the living in retribution.<br>Sent 3: The researchers on the island are killed by the newly risen zombies , except for Jenny , the daughter of a scientist couple.<br>Sent 4: She escapes , protected by an enchanted necklace charm given to her by her mother shortly before her death …<br>…<br>Sent 13: Jenny and Chuck flee , the only survivors remaining.<br>Sent 14: They stumble upon the cave once again, where the zombies appear and attack |
| | | Question: | Where did Chuck find weapons? |
| | | Candidates: | A. From the previous research team<br>B. Weapons were left behind by the long dead research team<br>C. Old research facilities medical quarters |
| | | Answer: | A and B |
| | | Question 2: | Who arms themselves against the zombies? |
| | | Candidates: | A. Jenny, the mercenaries, and Chuck<br>B. Chuck, Jenny and mercenaries<br>C. Defense<br>D. The hikers |
| | | Answer: | A and B and C |

We have created a statistical chart (Fig. 9) to visualize the distribution of various dataset styles presented in Table 1. Observing the corpus style, it's notable that textual tasks still constitute a substantial 93%. The proportion of multi-modal datasets remains relatively small, approximately 7%, underscoring the persisting challenges in the field of multi-modal MRC for future research. Delving into the complexity of the datasets, it becomes evident that the latest MRC datasets pose significant challenges, demanding a diverse set of reasoning skills. Approximately 13% of the datasets require logical reasoning for navigating well-structured arguments, while 20% necessitate commonsense reasoning, drawing upon general knowledge and contextual understanding beyond explicit information. Additionally, 10% of the datasets involve multi-hop reasoning, and 30% pose challenges in terms of complex reasoning. Examining the datasets in terms of domain, we find that 76% are in the open domain, with the remaining 24% in the closed domain. Regarding context style, datasets with a single paragraph take the lead at 59%, surpassing multi-paragraph datasets, which account for 31%. Shifting to question style, the natural form dominates at 59%, followed by the cloze type at 38%. Lastly, in terms of answer style, static answer candidates dominate at 62%, outnumbering datasets with dynamic answer candidates (28%) and entity candidates (10%). These insights collectively illuminate the intricate landscape of MRC datasets, emphasizing the varied challenges and considerations in their analysis and interpretation.

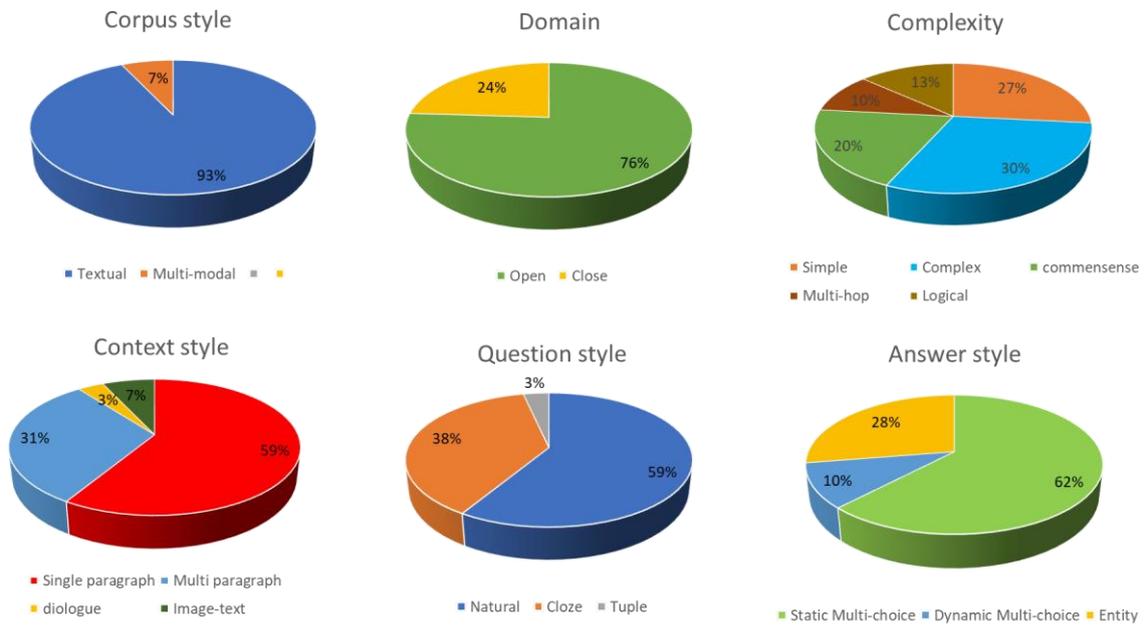

**Fig. 9.** Distribution of various dataset styles presented in Table 1.

## 3. Methodologies

The recent advancements in multi-choice machine reading comprehension have given rise to an array of methodologies that have revolutionized the field. These methodologies, while diverse in their approach, can be broadly categorized into two subsections: Fine-tuned Methods, and Prompt-tuned Methods. Fig. 10 shows a diagram comparing fine-tuned and prompt-tuned methods. Fine-tuning is a technique for adapting a pre-trained language model (PLM) to a specific task by retraining it on a large dataset. While prompt-tuning is a newer approach that is becoming increasingly popular. In prompt-tuning, the PLM is not trained on a large dataset. Instead, it is given a prompt, that provides instructions or context for the task. The PLM then uses its existing knowledge to generate a response.

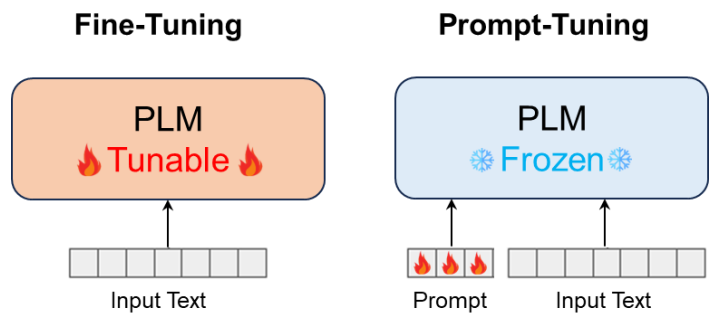

**Fig. 10.** Comparing fine-tuned and prompt-tuned methods.

## 3.1 Fine-tuned Methods

Fine-tuned methods mark a significant advancement in the field of MRC. These methods work by starting with a pretrained model, which has been trained on a large language corpus and thus, has a comprehensive understanding of language patterns, syntax, semantics, and contextual relationships. This pretrained model is then fine-tuned using a domain-specific dataset tailored to the target task. Fine-tuning involves adjusting the model parameters to optimize its performance for the task. This process enables the model to understand the subtle nuances and complexities of the task, allowing it to provide more accurate predictions. Refer to Fig. 11 for an illustrative depiction of these models. Fine-tuned methods involve integrating a PLM with a few straightforward linear layers, tailored to the specific task. This augmented model is then fine-tuned on the task-specific dataset. PLMs constitute the foundational layer of MRC. These methodologies leverage a wealth of pre-existing models that have been trained on vast amounts of unlabeled text data. The advantage of pretrained methods lies in their ability to handle complex language patterns and their robustness to a wide array of tasks without requiring task-specific modifications. Examples of PLMs are BERT (Devlin et al., 2018), GPT (OpenAI et al., 2023), XLNET (Yang et al., 2019), ALBERT (Lan et al., 2019) and RoBERTa (Y. Liu et al., 2019). They are typically employed as initial models, providing a solid base for further refinements and adaptations in the field of machine reading comprehension.

Most fine-tuned language models utilize the transformer architecture (Vaswani et al., 2017) for efficient processing of input sequences. Despite their effectiveness in modeling long-range dependencies in sequential data, traditional Transformers exhibit quadratic computational complexity and memory requirements, limiting their applicability to lengthy sequences. To address these scalability limitations, researchers have introduced various techniques, including sparse attention mechanisms. These approaches aim to reduce the computational burden and memory usage by selectively connecting certain elements within the self-attention mechanism of transformer layers. One notable approach is the Sparse Transformer (Child et al., 2019), which employs predefined sparse patterns, such as local or stride attention, to optimize computational efficiency. Other researchers have emphasized the importance of both local and global attention mechanisms (Ainslie et al., 2020; Beltagy et al., 2020; R. He et al., 2020; Zaheer et al., 2020). Local attention plays a crucial role in crafting content representations, while global attention enables the model to generate comprehensive sequence representations for predictive modeling. The Longformer (Beltagy et al., 2020) introduces a self-attention mechanism with a local window, considering both preceding and succeeding tokens for attention computation. Additionally, it incorporates global attention directed towards multiple predefined input tokens, particularly effective for learning task-specific representations. For instance, it can focus on the CLS token for classification tasks or question tokens for machine reading comprehension (MRC) tasks. The ETC model (Ainslie et al., 2020) enhances the transformer architecture by not only defining representations for input tokens but also introducing additional contextual representations as global tokens. Similarly, Big-Bird (Zaheer et al., 2020) augments the local-global attention structure of ETC by introducing scattered random attention patterns, allowing the model to consider information from various segments of the input sequence. The Realformer model (R. He et al., 2020) expands upon the ETC model by incorporating a residual attention layer. Furthermore, Jia et al. (2022) adopt a unique approach by extracting keywords from the text to guide the model's focus towards crucial details. However, this approach faces the challenge of conveying coherent meaning when individual keywords are considered independently.

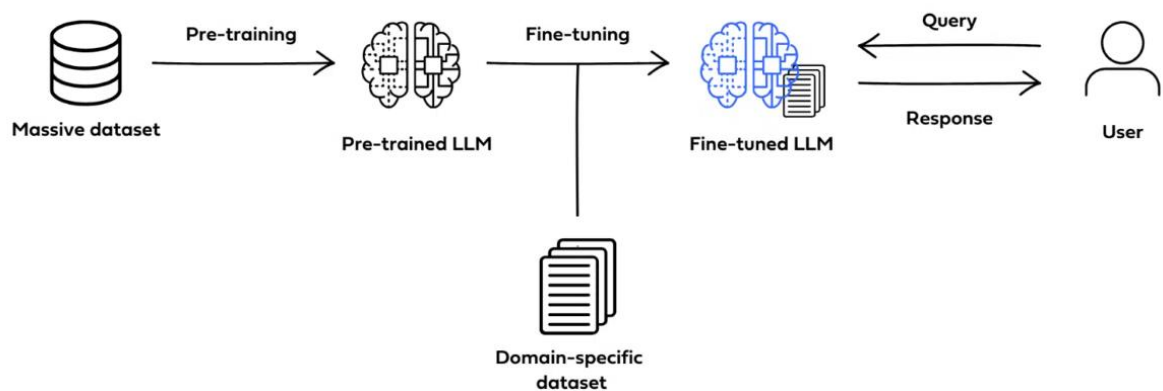

**Fig. 11.** Diagram of fine-tuned methods

Numerous researchers (Foolad & Kiani, 2023b; R. Li et al., 2021; Meng et al., 2023; Peng et al., 2021; Peters et al., 2019; Yamada et al., 2020; Z. Zhang et al., 2019) have explored methods to enhance PLMs by incorporating external knowledge sources. Li et al. (2021) utilized precomputed embeddings from ConceptNet (R. Li et al., 2021) as external knowledge representations, integrating them into BERT in three distinct configurations during the fine-tuning process. Additionally, they introduced a token-level search mechanism among multipartite relationships to filter external knowledge. Meng et al. (2023) proposed a confidence-based knowledge integration module that determines the amount of knowledge to incorporate based on confidence scores. However, external knowledge sources may not always be readily available, and their injection into models necessitates additional processing steps. This can increase the complexity of the model and make it more challenging to train and deploy. Moreover, it limits the model's ability to select the most relevant entities from the knowledge base, especially for ambiguous words that can introduce knowledge noise if not carefully chosen. Additionally, selecting the most relevant subgraphs from external knowledge and utilizing their information remains a challenge. To address these limitations, researchers have developed methods to inject knowledge base information into transformer models during the pre-training phase. The ERNIE (Z. Zhang et al., 2019) and Know-BERT (Peters et al., 2019) models exemplify this approach, learning fixed entity representations from a knowledge base. In contrast, the LUKE model (Yamada et al., 2020) employs a novel pre-training task to learn entity representations. The LUKE model extends RoBERTa by appending entities to the input of the transformer model and treating them as independent tokens. This enables the model to utilize the entity-aware self-attention mechanism to explicitly capture connections between entities. In other words, the LUKE model incorporates entity information into the self-awareness mechanism, enabling the model to focus on specific entities in the input text, thereby improving performance in tasks that require understanding entity relationships. Multi-hop MRC is a complex task that demands extensive understanding and logical connections between pieces of information to identify the correct response. Graph-based techniques (Foolad & Kiani, 2023a, 2023b; Mohammadi et al., 2023) simplify multi-hop MRC by constructing graphs. Luke-Graph (Foolad & Kiani, 2023b) enhances the LUKE model by eliminating irrelevant relationships and misleading data through the creation of a separate graph module. Foolad et al. (2023) propose an alternative approach, termed GESA, which integrates graph information into the attention mechanism of LUKE Transformer layers by introducing a novel attention pattern. This integration allows the attention mechanism to focus more on connected signals within the graph compared to unrelated ones, thus contributing to more accurate decision-making. Mohammadi et al. (2023) utilize more pertinent information from the context, including sentence topics, relationship topics, and the relative importance and strength of relationships, to filter paragraphs and construct the graph.

Recently, researchers (Chowdhery et al., 2022; P. He et al., 2020; Raffel et al., 2019) began exploring techniques to scale up the size and complexity of PLMs. This led to the emergence of large language models (LLMs), which are characterized by their massive parameter count, typically ranging from hundreds of billions to trillions. These models, including DeBERTa (P. He et al., 2020), T5 (Raffel et al., 2019), and PaLM (Chowdhery et al., 2022), have significantly expanded the capabilities of MRC. DeBERTa, standing for Decoding-enhanced BERT with disentangled attention, introduces a novel disentangled attention mechanism and a decoding-enhanced training procedure to enhance the efficiency and effectiveness of BERT. The disentangled attention mechanism in DeBERTa mitigates the impact of irrelevant information and amplifies the importance of relevant information in the attention process. Additionally, the decoding-enhanced training procedure guides the model to generate the output sequence in a left-to-right manner, rather than in a random order, further improving performance. In contrast to DeBERTa, T5 adopts a text-to-text transformer architecture and is pre-trained on a diverse range of tasks. This versatility allows T5 to perform a wide spectrum of NLP tasks by converting them into a text-to-text format. PaLM, short for Pathways-augmented Language Model, innovates by combining traditional transformer-based models with a novel pathways mechanism. This mechanism enables more efficient processing of long sequences by dividing the input sequence into smaller segments and processing each segment separately.

Table 5 provides a comprehensive comparison of the most widely used PLMs in MRC. The table includes key information such as the model's release year, the number of parameters, input length capacity, underlying Transformer architecture (whether it employs an encoder, decoder, or both), as well as their respective benefits and challenges. Notable models like BERT and RoBERTa stand out for their widespread adoption and superior context understanding. On the other hand, models like XLNET excel in scenarios where bidirectional context is crucial, while ALBERT employs parameter reduction techniques. Know-BERT and ERNIE3.0 incorporates external knowledge bases, but necessitates a large knowledge graph. Longformer and ETC are adept at handling long documents, yet the former faces memory consumption challenges, and the latter exhibits complex model architecture. Large scale of GPT-3 enables profound understanding and few-shot learning, but requires substantial computational resources. unique ability of T5 to transform diverse tasks into a text-to-text format is counterbalanced by its computational requirements. Lastly, DeBERTa and PaLM showcase advanced reasoning capabilities, demanding significant computational resources. This comparative analysis aids in understanding the strengths and limitations of each PLM, facilitating informed model selection based on specific MRC requirements.

Table 6 provides a comprehensive overview of the performance of various Pre-trained Language Models (PLMs) across MRC datasets, employing F1/EM metrics for ReCoRD and MultiRC, and accuracy (ACC) for other datasets. The human baseline is included for reference.

Notably, T5, DeBERTa, PaLM, and ERNIE3.0 consistently demonstrate high accuracy across multiple datasets, outperforming both BERT and RoBERTa. T5 achieves top scores across most evaluated datasets such as ARC, CosmosQA, and QASC, while DeBERTa excels in ReClor. Moreover, PaLM and ERNIE3.0 perform well in ReCoRD and MultiRC Overall, the results showcase that the performance of PLMs in MRC tasks has continued to improve significantly. RoBERTa, T5, DeBERTa, PaLM, and ERNIE3.0 consistently achieve state-of-the-art performance across a variety of MRC datasets.

**Table 5.** Comparing the most used PLMs in MRC.

| Model | Year | # Parameters | Input length | Transformer architecture | Benefits | Challenges |
|---|---|---|---|---|---|---|
| **BERT** (Devlin et al., 2018) | 2018 | 340M | 512 | Encoder | Widely used model, Superior context understanding | Difficulty with long documents |
| **RoBERTa** (Y. Liu et al., 2019) | 2019 | 355M | 512 | Encoder | Widely used model, improved training techniques | Same as BERT |
| **XLNET** (Yang et al., 2019) | 2019 | 340M | 512 | Decoder | Excels in scenarios where bidirectional context is crucial | Requires careful hyperparameter tuning |
| **ALBERT** (Lan et al., 2019) | 2019 | 18M | 512 | Encoder | Parameter reduction techniques | Same as BERT |
| **Know-BERT** (Peters et al., 2019) | 2019 | 516M | 512 | Encoder | Incorporates external knowledge bases | Requires a large knowledge graph |
| **Longformer** (Beltagy et al., 2020) | 2020 | 102M | 4096 | Encoder | Handles long documents | Memory consumption |
| **ETC** (Ainslie et al., 2020) | 2020 | 558M | 4096 | Encoder | Scaling input length and encoding structured inputs | Complex model architecture |
| **LUKE** (Yamada et al., 2020) | 2020 | 483M | 512 | Encoder | Enhanced named entity recognition | Complexity of entity embeddings |
| **GPT-3** (Brown et al., 2020) | 2020 | 175B | 4096 | Encoder-Decoder | Large Scale Understanding, Few-Shot Learning | Computational Resources |
| **T5** (Raffel et al., 2019) | 2019 | 11B | 2048 | Encoder-Decoder | Ability to transform different tasks into a text-to-text format | Computational requirements |
| **DeBERTa** (P. He et al., 2020) | 2022 | 1.5B | 24,528 | Decoder | High efficiency in reasoning tasks, Handling long sequence input | Scalability issues |
| **PaLM** (Chowdhery et al., 2022) | 2022 | 540B | 2048 | Encoder | Capabilities in advanced reasoning | Computational Resources |
| **ERNIE3.0** (S. Wang et al., 2021) | 2022 | 10B | 512 | Encoder | Incorporating external knowledge | Its dependency on external knowledge sources |

**Table 6.** the performance of various PLMs across the MRC datasets.

| Model | ReCoRD (F1/EM) | MultiRC (F1/EM) | ARC (ACC) | DREAM (ACC) | CosmosQA (ACC) | QASC (ACC) | ReClor (ACC) |
|---|---|---|---|---|---|---|---|
| **Human** | *91.7/91.3* | *81.8/51.9* | - | *98.6* | *94.0* | *93.0* | *63.0* |
| **BERT** (Devlin et al., 2018) | 72.0/71.3 | 70.0/24.1 | 44.6 | 66.8 | 67.3 | 53.1 | 49.8 |
| **RoBERTa** (Y. Liu et al., 2019) | 90.6/90.0 | 84.4/52.5 | 66.4 | 88.9 | 80.8 | 80.0 | 55.6 |
| **XLNET** (Yang et al., 2019) | 82.7/81.5 | - | 67.0 | 72.0 | - | - | 56.0 |
| **ALBERT** (Lan et al., 2019) | 84.3/83.5 | 75.9/35.1 | 62.9 | **90.0** | 79.2 | - | 62.6 |
| **T5** (Raffel et al., 2019) | 94.1/93.4 | 88.1/63.3 | *81.1* | - | *90.2* | *89.5* | - |
| **DeBERTa** (P. He et al., 2020) | 94.5/94.1 | 88.2/63.7 | - | - | 86.8 | 89.3 | *72.7* |
| **PaLM** (Chowdhery et al., 2022) | 94.2/93.3 | *88.7/63.6* | - | - | - | - | - |
| **ERNIE3.0** (S. Wang et al., 2021) | *94.7/94.2* | 88.6/63.2 | - | - | - | - | - |

### 3.2 Prompt-tuned Methods

With remarkable strides in developing pre-trained language models, as illustrated in Fig. 12, large language models (LLMs) are progressively emerging with augmented parameters and enhanced learning capabilities. As of January 1, 2024, the most expansive model, Gemini (Team et al., 2023), crafted by Google, boasts an impressive eight trillion parameters. In essence, the large number of parameters allows the model to learn more complex patterns. However, training these larger language models demands substantial data and heightened computational resources, posing a significant challenge. This challenge can hinder the deployment of models on devices with constrained resources. Additionally, dealing with the expanding scale of LLM parameters necessitates a more resource-intensive fine-tuning process. For instance, deploying an LLM model with 175 billion parameters requires a minimum of 350 gigabytes of GPU memory, leveraging specialized architectures. Unfortunately, contemporary LLMs with trillions of parameters impose even more substantial demands on memory and computational power. These computational requirements often exceed the practical capabilities of the majority of product teams. Mitigating this computational strain can be achieved through the adoption of prompt learning techniques (Brown et al., 2020; X. L. Li & Liang, 2021; P. Liu et al., 2023).

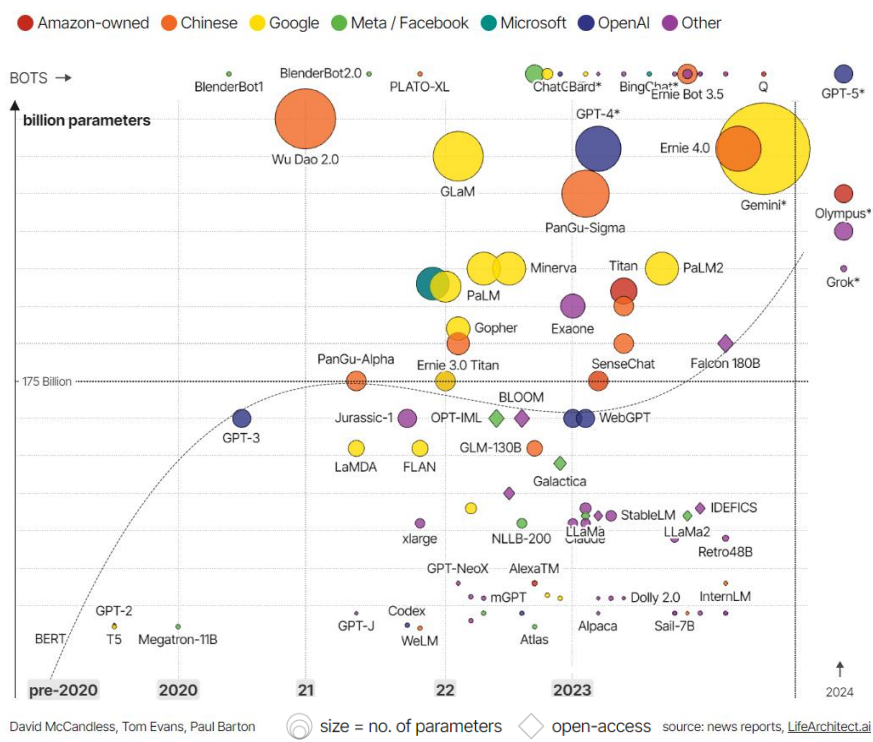

**Fig. 12.** Comparing the size of PLMs

Following the success of GPT-3 (Brown et al., 2020), prompt learning (P. Liu et al., 2023)

has presented another efficient solution for utilizing PLMs. Due to its low computational cost and high accuracy, prompt learning has garnered widespread attention. In prompt learning, there is a need for a prompt that can either be a human-readable text (discrete prompts (Brown et al., 2020; Min et al., 2022; Wei et al., 2021, 2022)) or an embedded vector (continuous prompts (Chen et al., 2023; Clive et al., 2021; Lester et al., 2021; X. L. Li & Liang, 2021; Ma et al., 2022; Tang et al., 2022)). Table 7 shows a comparison of discrete and continuous prompts. Discrete prompts are the simplest type of prompt that provides instructions or context for the task. Discrete prompts are easy to understand and use, but they can be limited in their flexibility. For example, it can be difficult to use discrete prompts to express complex relationships or concepts. While continuous prompts are more flexible than discrete prompts because they use embedding vectors instead of text. Embedding vectors are numerical representations of words and phrases that can capture subtle semantic relationships. This makes continuous prompts more powerful for expressing complex tasks, but they can also be more difficult to understand and use. Subsequently, we will discuss the types of discrete and continuous prompts shown in Fig. 13.

**Table 7.** comparison of discrete and continuous prompts

| Feature | Discrete Prompts | Continuous Prompts |
|---|---|---|
| **Format** | Human-readable text | Embedding vectors |
| **Flexibility** | Less flexible | More flexible |
| **Training** | Not typically trained | Typically trained |
| **Applications** | Simpler tasks | More complex tasks |

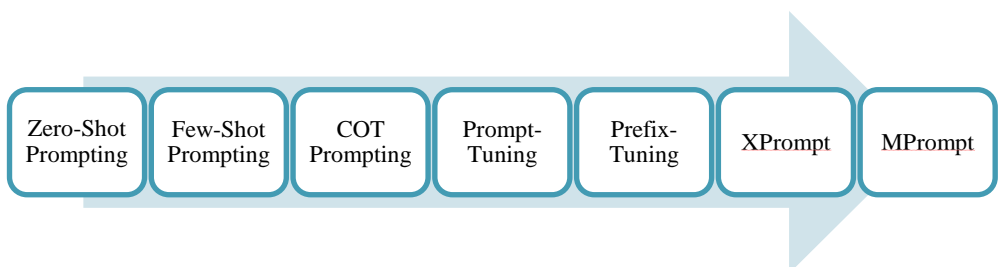

**Fig. 13.** Types of discrete and continuous prompts

Since LLMs have been extensively trained on a large amount of data, they possess the ability to perform tasks without specific training, known as zero-shot or few-shot learning, as illustrated in Fig. 14. Zero-shot learning performs tasks without any labeled data and any direct examples for that task. Instead, the model is given a prompt, which is a short piece of text that provides instructions or context for the task. Wei et al. (2021) improved zero-shot learning by introducing the instruction-tuning method. In this approach, fine-tuning the models is done through instruction, followed by Reinforcement Learning with Human Feedback (RLHF) (Paulus et al., 2017), aligning the model more effectively with human preferences (similar to ChatGPT). However, for more complex tasks, zero-shot learning falls short. Addressing this limitation, few-shot learning methods (Brown et al., 2020; Min et al., 2022) have been introduced, utilizing one or a few examples as instances within prompts for the intended task. Nevertheless, these methodologies prove less effective for more complex reasoning tasks. Hence, advanced approaches such as chain-of-thought (COT) (Wei et al., 2022) have been proposed. In COT prompting, each example in the prompt is accompanied by a logical rationale for the response, offering a detailed step-by-step explanation of problem-solving and enriching the model's understanding of the task. Hsieh and et al. (2023) applied the COT approach to generate explanatory labels alongside logical rationales using LLMs. Subsequently, these logical rationales, coupled with task labels, were employed to train smaller models tailored to specific tasks. In essence, these rationales provide a more nuanced and information-rich understanding of why a particular input correlates with a specific output, often encapsulating task-related knowledge. Table 8 shows a prompt sample of MRC task for each discrete prompt technique.

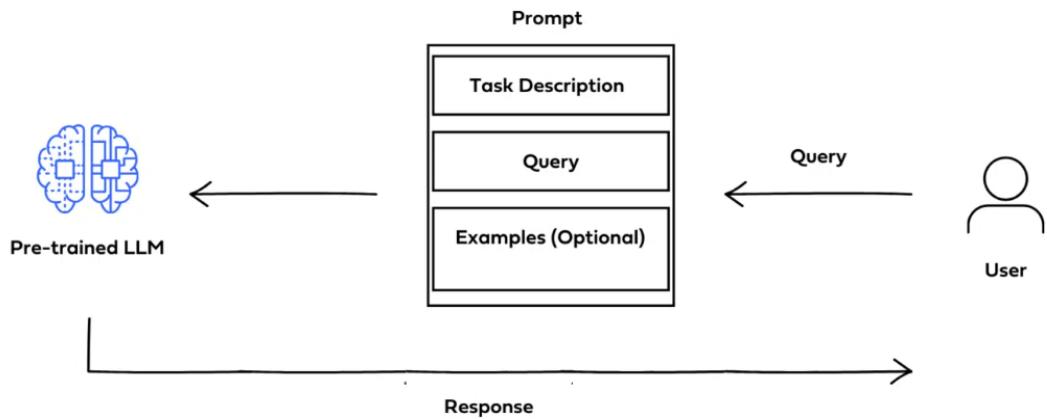

**Fig. 14.** Diagram of prompt-tuned methods

**Table 8.** prompt sample of MRC task for each discrete prompt technique

| Prompt Technique | Prompt |
|---|---|
| **Zero-Shot Prompting** (Wei et al., 2021) | Answer to the given question from the provided options based on the given context. <br> *Context:* .... A large enough comet colliding with Earth could have caused a cloud of dust that enshrouded the planet and cooled the climate long enough to result in the dinosaurs' demise. <br> *Question:* Which one of the following statements, most seriously weakens the argument? <br> *Options:* <br> A. Many other animal species from same era did not become extinct at the same time the dinosaurs did. <br> B. It cannot be determined from dinosaur skeletons whether the animals died from the effects of a dust cloud. <br> C. The consequences for vegetation and animals of a comet colliding with Earth are not fully understood. <br> D. Various species of animals from the same era and similar to them in habitat and physiology did not become extinct. <br> *Answer:* |
| **Few-Shot Prompting** (Brown et al., 2020; Min et al., 2022) | Answer to the given question from the provided options based on the given context. <br> *Example 1:* <br> --------------------------------------------- <br> *Context:* In jurisdictions where use of headlights is optional when ... <br> *Question:* Which one of the following, if true, most helps to resolve the apparent discrepancy in the information above? <br> *Options:* <br> A. In jurisdictions where use of headlights is optional when visibility is good, one ... <br> B. Only very careful drivers use headlights when their use is not legally required. <br> C. The jurisdictions where use of headlights is mandatory at all times are those where … <br> *Answer:* B <br> --------------------------------------------- <br> *Context:* .... A large enough comet colliding with Earth could have caused a cloud of dust that |

| | enshrouded the planet and cooled the climate long enough to result in the dinosaurs' demise.<br>*Question:* Which one of the following statements, most seriously weakens the argument?<br>*Options:*<br>A. Many other animal species from same era did not become extinct at the same time the dinosaurs did.<br>B. It cannot be determined from dinosaur skeletons whether the animals died from the effects of a …<br>C. The consequences for vegetation and animals of a comet colliding with Earth are not fully …<br>D. D. Various species of animals from the same era and similar to them in habitat and physiology …<br>*Answer:* |
|---|---|
| **COT Prompting**<br>(Wei et al., 2022) | Answer to the given question from the provided options based on the given context.<br>*Example 1:*<br>---------------------------------------------<br>*Context:* In jurisdictions where use of headlights is optional when …<br>*Question:* Which one of the following, if true, most helps to resolve the apparent discrepancy in the information above?<br>*Options:*<br>A. In jurisdictions where use of headlights is optional when visibility is good, one …<br>B. Only very careful drivers use headlights when their use is not legally required.<br>C. The jurisdictions where use of headlights is mandatory at all times are those where …<br>***Rationale** with Answer: option B provides a logical connection between the observed behavior of drivers who use headlights voluntarily and the lack of collision reduction when headlights are made mandatory for all, thereby resolving the apparent contradiction in the given information ….* **The answer is B.**<br>---------------------------------------------<br>*Context:* …. A large enough comet colliding with Earth could have caused a cloud of dust that enshrouded the planet and cooled the climate long enough to result in the dinosaurs' demise.<br>*Question:* Which one of the following statements, most seriously weakens the argument?<br>*Options:*<br>A. Many other animal species from same era did not become extinct at the same time the dinosaurs did.<br>B. It cannot be determined from dinosaur skeletons whether the animals died from the effects of a …<br>C. The consequences for vegetation and animals of a comet colliding with Earth are not fully …<br>D. Various species of animals from the same era and similar to them in habitat and physiology did …<br>***Rationale** with Answer:* |

The continuous prompts introduce a more flexible solution, encoding information in an embedding vector format to efficiently present them in pre-trained models. For example, Li and Liang (2021) proposed prefix-tuning approach where an optimized sequence of prefixes is added to each layer of a transformer while keeping PLM parameters constant. This technique serves as a lightweight alternative for fine-tuning and has demonstrated comparable performance with fewer trainable parameters. Another noteworthy contribution comes from Lester et al (2021), who advocate for prompt-tuning. This method incorporates trainable prompts into the input and employs fewer parameters compared to prefix-tuning. However, Ma et al. (2022) discovered negative cues in prompt-tuning that adversely affect NLP task performance. They introduced XPrompt to address these negative cues, leading to performance improvement.

While the aforementioned methods are innovative, they are not tailored to specific inputs, using a consistent prompt for all inputs of a given task and not fully leveraging input semantics to generate responses. To solve this limitation, Tang et al. (2022) devised a strategy to extract content-based prompts from external PLMs based on input text, resulting in superior performance in natural language generation. Going a step further, Clive et al. (2021) ingeniously fused task-specific prompts with dynamic prompts, providing the model with more precise control over the generated text. Another standout model in this realm is MPrompt (Clive et al., 2021), which introduces prompts at multiple levels for task-specific, domain-specific, and context-specific purposes. This approach not only enhances the textual understanding of PLMs but also significantly improves MRC task performance. In summary, Table 9 presents an exhaustive comparison of various prompting techniques.

Table 9. comparing prompting techniques for PLMs.

| Prompting Technique | Benefits | Challenges |
|---|---|---|
| **Zero-Shot Prompting** (Wei et al., 2021) | Efficient and scalable for new tasks, requires little or no labeled data | Can be limited in accuracy for complex tasks |
| **Few-Shot Prompting** (Brown et al., 2020; Min et al., 2022) | Offers improved performance compared to zero-shot prompting, using a few examples as instances | Still requires some labeled data and may not be effective for highly complex tasks |
| **COT Prompting** (Wei et al., 2022) | Provides more detailed explanations and rationales for task performance, leading to better understanding and generalization | Can be more complex to implement and may not be suitable for all tasks |
| **Prompt-Tuning** (Lester et al., 2021) | More efficient than fine-tuning, using trainable prompts instead of extensive parameter adjustments | Can still be computationally expensive for certain tasks |
| **Prefix-Tuning** (X. L. Li & Liang, 2021) | Lightweight alternative to fine-tuning, using optimized prefixes to modify model input | Offers lower performance compared to prompt-tuning |
| **XPrompt** (Ma et al., 2022) | Addressing negative cues in prompt-tuning, leading to improved performance | Requires more careful design of prompts to avoid introducing new biases |
| **MPrompt** (Chen et al., 2023) | Incorporating prompts at multiple levels, enhancing textual understanding and MRC task performance | Can be more complex to implement and may not be suitable for all tasks |

## 4. Challenges

Despite the remarkable progress achieved in MRC through fine-tuned and prompt-tuned methods, several challenges remain. These challenges hinder the further development and applicability of these methods in real-world scenarios.

- **Data Scarcity and Bias:** The availability of high-quality MRC datasets is crucial for training and evaluating models. However, existing datasets often suffer from limitations in size, diversity, and representativeness of real-world reading comprehension scenarios. This data scarcity can lead to overfitting, limited generalization capabilities, and potential biases embedded in the datasets. Moreover, fine-tuning multiple in-domain datasets simultaneously can lead to improvements, but it may also result in performance drops on easier datasets (Pan et al., 2019).
- **Complex Reasoning:** MRC tasks often require complex deductive, inductive, or abductive reasoning, which may be beyond the capabilities of current models. This includes tasks that demand understanding causal relationships, drawing inferences from incomplete or ambiguous information, or applying knowledge from one domain to another.
- **Commonsense Knowledge Utilization:** Effective comprehension often requires access to commonsense knowledge, which is not explicitly provided in the text or question. While some models incorporate external knowledge sources, their ability to leverage this knowledge effectively remains limited.
- **Explanation and Interpretability:** The inner workings of deep learning models, including fine-tuned and prompt-tuned models, can be highly complex and non-transparent. This lack of interpretability makes it difficult to understand the reasoning behind model decisions, hindering error analysis and debugging.
- **Domain Adaptation and Cross-Lingual Transfer:** Despite their impressive performance

on specific datasets, fine-tuned and prompt-tuned models often struggle to generalize to new domains or languages. Datasets in different domains may have distinct characteristics, and languages may exhibit linguistic variations that require adaptation.
- **Computational Cost and Scalability:** Fine-tuning large pre-trained language models (PLMs) for MRC tasks is computationally expensive and requires substantial memory resources. This is because the entire model needs to be updated with new data, which can be resource-intensive. This computational burden can limit their practical applications, especially in resource-constrained environments.
- **Robustness to Noise and Errors:** MRC tasks are susceptible to noise and errors in the text, questions, or answers. Existing models may not be robust to these irregularities, leading to incorrect predictions.
- **Handling Unseen Answer Types and Ambiguous Questions:** MRC models often perform well on questions with well-defined answer types and clear contexts. However, they may struggle with questions that present unexpected answer formats or ambiguous contexts, requiring more sophisticated reasoning and generalization capabilities.

By addressing these challenges, we can further advance the field of MRC and develop models that can effectively and reliably answer complex questions from text, enabling a wider range of applications in various domains.

## 5. Future Direction

The field of MRC has witnessed remarkable advancements in recent years, with the development of fine-tuned and prompt-tuned methods significantly improving performance on various benchmarks. However, several challenges remain, and there is ample opportunity for further research and development.
- **Enhanced Data Representation and Augmentation:** Expanding the size and diversity of MRC datasets is crucial for enabling models to generalize better to real-world scenarios. This includes creating more comprehensive datasets that encompass a wider range of linguistic styles, genres, and domains. Additionally, data augmentation techniques can be employed to artificially generate new examples, enriching the training data and reducing data scarcity.
- **Addressing Complex Reasoning and Real-World Challenges:** MRC tasks often require complex deductive, inductive, or abductive reasoning abilities, which may go beyond the capabilities of current models. To address this, research should focus on developing models that can perform complex reasoning tasks, such as understanding causal relationships, drawing inferences from incomplete or ambiguous information, or applying knowledge from one domain to another.
- **Incorporating Commonsense Knowledge and Reasoning:** Effective comprehension often requires access to commonsense knowledge, which is not explicitly provided in the text or question. To address this, researchers should investigate methods for integrating commonsense knowledge sources into MRC models. This could involve incorporating knowledge graphs, concept ontologies, or other knowledge repositories to enable models to draw inferences and reason over complex relationships.
- **Improving Explainability and Interpretability:** The opacity of deep learning models hinders understanding the reasoning behind their decisions, making it challenging to debug errors and assess their trustworthiness. To address this, research should focus on developing explainable MRC models that can provide insights into their decision-making processes. This could involve techniques such as attention maps, attribution methods, or model introspection capabilities.

- **Enhancing Domain Adaptation and Cross-Lingual Transferability**: Current models often struggle to generalize to new domains or languages, requiring adaptation or retraining. To address this, research should explore methods for improving domain adaptation and cross-lingual transferability. This could involve developing domain-specific representations, utilizing multilingual embeddings, or exploring task-oriented learning approaches.
- **Optimizing Computational Efficiency and Scalability:** Large language models used in prompt-tuned approaches can be computationally expensive and time-consuming, limiting their practical applications. To address this, research should focus on developing efficient and scalable MRC models that can operate on resource-constrained devices. This could involve techniques such as model compression, knowledge distillation, or hardware acceleration. Delta-tuning methods have been proposed as a more parameter-efficient fine-tuning (PEFT) of PLMs, which update only a small number of parameters, potentially reducing computational and storage costs (Ding et al., 2023). Moreover, prompt-tuning techniques offer a more efficient and lightweight approach.
- **Handling Noisy Data:** MRC models are susceptible to noise and errors in the text, questions, or answers. To improve robustness, research should explore methods for handling noisy and errorful data. This could involve techniques for data cleaning, error correction, or robust learning algorithms.
- **Addressing Ambiguous and Unseen Answer Types:** MRC models often perform well on questions with well-defined answer types and clear contexts. However, they may struggle with questions that present unexpected answer formats or ambiguous contexts. To address this, research should focus on developing models that can effectively handle ambiguous and unseen answer types, requiring more sophisticated reasoning and generalization capabilities.

By addressing these challenges and exploring these future directions, the field of MRC can continue to advance, leading to the development of more powerful and versatile models that can effectively answer complex questions from text, enabling a wider range of applications in various domains. As research in MRC continues to advance, we can expect further breakthroughs that will bring us closer to human-level reading comprehension.

## 6. Conclusion

Machine reading comprehension (MRC) has undergone remarkable advancements in recent years, particularly in the realm of multi-choice and cloze-style datasets. Cloze-style MRC tasks involve filling in the blanks in a given sentence or passage, while multi-choice MRC requires selecting the correct answer from a set of options. Benchmark datasets and methodologies have emerged to assess and refine MRC models' performance in both these domains. The insights provided in this survey serve as a valuable guide for researchers and practitioners in their future endeavors in MRC.

However, our comprehensive survey reveals that while some MRC datasets receive substantial attention from researchers, others remain relatively obscure within the community. To bridge this gap, our study provides an in-depth analysis of 30 existing cloze style and multiple-choice MRC benchmark datasets. Our findings underscore the need for further research to bridge the divide between highly scrutinized and overlooked datasets within the MRC community. We summarized the characteristics of each dataset, including data sources, evaluation metrics, SOTA model performances, and solved status. Additionally, we proposed a refined classification method for cloze-style and multi-choice MRC datasets based on corpus style, domain, complexity, context style, question style, and answer style. This refined classification approach offers a more comprehensive understanding of the diverse attributes of each dataset. Notably, it categorizes the datasets according to their complexity, a crucial aspect of recent datasets.

Furthermore, we conducted a comprehensive exploration of the latest methodologies and state-of-the-art models in the realm of MRC, focusing on fine-tuned and prompt-tuned methods. Fine-tuned methods, which involve training pre-trained language models (PLMs) on task-specific data, have become the dominant approach. However, prompt-tuned methods, which utilize prompts to guide the PLM's response generation, have recently gained traction due to their lower computational overhead and ability to handle zero-shot or few-shot learning scenarios.

Despite the significant progress in MRC, several persistent challenges remain, as we have thoroughly investigated in this survey. Future research should prioritize addressing these challenges and pushing the boundaries of machine comprehension and response capabilities. We have also identified open research directions and potential future directions in this field. The progress made in recent years in MRC has shown great promise, and we expect further advances in this field in the coming years.

## References

Ainslie, J., Ontañón, S., Alberti, C., Cvicek, V., Fisher, Z., Pham, P., Ravula, A., Sanghai, S., Wang, Q., & Yang, L. (2020). ETC: Encoding Long and Structured Inputs in Transformers. *EMNLP 2020 - 2020 Conference on Empirical Methods in Natural Language Processing, Proceedings of the Conference*, 268–284. https://doi.org/10.18653/V1/2020.EMNLP-MAIN.19

Bajgar, O., Kadlec, R., & Kleindienst, J. (2016). *Embracing data abundance: BookTest Dataset for Reading Comprehension*. https://arxiv.org/abs/1610.00956v1

Baradaran, R., Ghiasi, R., & Amirkhani, H. (2020). A Survey on Machine Reading Comprehension Systems. *Natural Language Engineering*, *28*(6), 683–732. https://doi.org/10.1017/S1351324921000395

Beltagy, I., Peters, M. E., & Cohan, A. (2020). *Longformer: The Long-Document Transformer*. https://doi.org/10.48550/arxiv.2004.05150

Brown, T. B., Mann, B., Ryder, N., Subbiah, M., Kaplan, J., Dhariwal, P., Neelakantan, A., Shyam, P., Sastry, G., Askell, A., Agarwal, S., Herbert-Voss, A., Krueger, G., Henighan, T., Child, R., Ramesh, A., Ziegler, D. M., Wu, J., Winter, C., … Amodei, D. (2020). Language Models are Few-Shot Learners. *Advances in Neural Information Processing Systems*, *2020-December*. https://arxiv.org/abs/2005.14165v4

Chaudhury, A., Tapaswi, M., Kim, S. W., & Fidler, S. (2019). *The Shmoop Corpus: A Dataset of Stories with Loosely Aligned Summaries*. https://arxiv.org/abs/1912.13082v2

Chen, G., Qian, Y., Wang, B., & Li, L. (2023). *MPrompt : Exploring Multi-level Prompt Tuning for Machine Reading Comprehension*. 5163–5175.

Child, R., Gray, S., Radford, A., & Sutskever, I. (2019). *Generating Long Sequences with Sparse Transformers*. https://doi.org/10.48550/arxiv.1904.10509

Chowdhery, A., Narang, S., Devlin, J., Bosma, M., Mishra, G., Roberts, A., Barham, P., Chung, H. W., Sutton, C., Gehrmann, S., Schuh, P., Shi, K., Tsvyashchenko, S., Maynez, J., Rao, A., Barnes, P., Tay, Y., Shazeer, N., Prabhakaran, V., … Fiedel, N. (2022). *PaLM: Scaling Language Modeling with Pathways*. https://arxiv.org/abs/2204.02311v5

Clark, P., Cowhey, I., Etzioni, O., Khot, T., Sabharwal, A., Schoenick, C., & Tafjord, O. (2018). Think you have Solved Question Answering? Try ARC, the AI2 Reasoning Challenge. *ArXiv:1803.05457*. http://arxiv.org/abs/1803.05457

Clive, J., Cao, K., & Rei, M. (2021). Control Prefixes for Parameter-Efficient Text Generation. *GEM 2022 - 2nd Workshop on Natural Language Generation, Evaluation, and Metrics, Proceedings of the Workshop*, 363–382. https://doi.org/10.18653/v1/2022.gem-1.31

Cui, Y., Liu, T., Che, W., Chen, Z., & Wang, S. (2021). ExpMRC: Explainability Evaluation for Machine Reading Comprehension. *Heliyon*, *8*(4). https://doi.org/10.1016/j.heliyon.2022.e09290

Devlin, J., Chang, M. W., Lee, K., & Toutanova, K. (2018). BERT: Pre-training of Deep Bidirectional Transformers for Language Understanding. *NAACL HLT 2019 - 2019 Conference of the North American Chapter of the Association for Computational Linguistics: Human Language Technologies - Proceedings of the Conference*, *1*, 4171–4186. https://doi.org/10.48550/arxiv.1810.04805

Dhingra, B., Mazaitis, K., & Cohen, W. W. (2017). *Quasar: Datasets for Question Answering by Search and Reading*. https://arxiv.org/abs/1707.03904v2

Ding, N., Qin, Y., Yang, G., Wei, F., Yang, Z., Su, Y., Hu, S., Chen, Y., Chan, C. M., Chen, W., Yi, J., Zhao, W., Wang, X., Liu, Z., Zheng, H. T., Chen, J., Liu, Y., Tang, J., Li, J., & Sun, M. (2023). Parameter-efficient fine-tuning of large-scale pre-trained language models. *Nature Machine Intelligence 2023 5:3*, *5*(3), 220–235. https://doi.org/10.1038/s42256-023-00626-4

Dzendzik, D., Vogel, C., & Foster, J. (2021). English Machine Reading Comprehension Datasets: A Survey. *EMNLP 2021 - 2021 Conference on Empirical Methods in Natural Language Processing, Proceedings*, 8784–8804. https://doi.org/10.18653/v1/2021.emnlp-main.693

Foolad, S., & Kiani, K. (2023a). Integrating a Heterogeneous Graph with Entity-aware Self-attention using Relative Position Labels for Reading Comprehension Model. *ArXiv Preprint ArXiv:2307.10443*.

Foolad, S., & Kiani, K. (2023b). LUKE-Graph: A Transformer-based Approach with Gated Relational Graph Attention for Cloze-style Reading Comprehension. *Neurocomputing*, *558*. https://doi.org/10.1016/j.neucom.2023.126786

He, P., Liu, X., Gao, J., Chen, W., & Dynamics, M. (2020). *DeBERTa: Decoding-enhanced BERT with Disentangled Attention*. https://arxiv.org/abs/2006.03654v6

He, R., Ravula, A., Kanagal, B., & Ainslie, J. (2020). RealFormer: Transformer Likes Residual Attention. *Findings of the Association for Computational Linguistics: ACL-IJCNLP 2021*, 929–943. https://doi.org/10.18653/v1/2021.findings-acl.81

Hill, F., Bordes, A., Chopra, S., & Weston, J. (2015). The goldilocks principle: Reading children's books with explicit memory representations. *ArXiv Preprint ArXiv:1511.02301*.

Hsieh, C. Y., Li, C. L., Yeh, C. K., Nakhost, H., Fujii, Y., Ratner, A., Krishna, R., Lee, C. Y., & Pfister, T. (2023). Distilling Step-by-Step! Outperforming Larger Language Models with Less Training Data and Smaller Model Sizes. *Proceedings of the Annual Meeting of the Association for Computational Linguistics*, 8003–8017. https://doi.org/10.18653/v1/2023.findings-acl.507

Huang, L., Le Bras, R., Bhagavatula, C., & Choi, Y. (2019). Cosmos QA: Machine Reading Comprehension with Contextual Commonsense Reasoning. *EMNLP-IJCNLP 2019 - 2019 Conference on Empirical Methods in Natural Language Processing and 9th International Joint Conference on Natural Language Processing, Proceedings of the Conference*, 2391–2401. https://doi.org/10.18653/V1/D19-1243

Jia, M., Liao, L., Wang, W., Li, F., Chen, Z., Li, J., & Huang, H. (2022). Keywords-aware dynamic graph neural network for multi-hop reading comprehension. *Neurocomputing*, *501*, 25–40. https://doi.org/10.1016/J.NEUCOM.2022.05.110

Jin, D., Pan, E., Oufattole, N., Weng, W. H., Fang, H., & Szolovits, P. (2020). What Disease does this Patient Have? A Large-scale Open Domain Question Answering Dataset from Medical Exams. *Applied Sciences (Switzerland)*, *11*(14). https://doi.org/10.3390/app11146421

Kawabata, A., & Sugawara, S. (2023). Evaluating the Rationale Understanding of Critical Reasoning in Logical Reading Comprehension. *ACL*. https://arxiv.org/abs/2311.18353v1

Khashabi, D., Chaturvedi, S., Roth, M., Upadhyay, S., & Roth, D. (2018). Looking Beyond the Surface:A Challenge Set for Reading Comprehension over Multiple Sentences. *Proceedings of North American Chapter of the Association for Computational Linguistics (NAACL)*.

Khot, T., Clark, P., Guerquin, M., Jansen, P., & Sabharwal, A. (2019). QASC: A Dataset for Question Answering via Sentence Composition. *AAAI 2020 - 34th AAAI Conference on Artificial Intelligence*, 8082–8090. https://doi.org/10.1609/aaai.v34i05.6319

Lai, G., Xie, Q., Liu, H., Yang, Y., & Hovy, E. (2017). Race: Large-scale reading comprehension dataset from examinations. *ArXiv Preprint ArXiv:1704.04683*.

Lan, Z., Chen, M., Goodman, S., Gimpel, K., Sharma, P., & Soricut, R. (2019). ALBERT: A Lite BERT for Self-supervised Learning of Language Representations. *8th International Conference on Learning Representations, ICLR 2020*. https://arxiv.org/abs/1909.11942v6

Lester, B., Al-Rfou, R., & Constant, N. (2021). The Power of Scale for Parameter-Efficient Prompt Tuning. *EMNLP 2021 - 2021 Conference on Empirical Methods in Natural Language Processing, Proceedings*, 3045–3059. https://doi.org/10.18653/v1/2021.emnlp-main.243

Li, R., Jiang, Z., Wang, L., Lu, X., Zhao, M., & Chen, D. (2021). Enhancing Transformer-based language models with commonsense representations for knowledge-driven machine comprehension. *Knowledge-Based Systems*, *220*, 106936. https://doi.org/10.1016/J.KNOSYS.2021.106936

Li, X. L., & Liang, P. (2021). Prefix-Tuning: Optimizing Continuous Prompts for Generation. *ACL-IJCNLP 2021 - 59th Annual Meeting of the Association for Computational Linguistics and the 11th International Joint Conference on Natural Language Processing, Proceedings of the Conference*, 4582–4597. https://doi.org/10.18653/v1/2021.acl-long.353

Liang, Y., Li, J., & Yin, J. (2019). A New Multi-choice Reading Comprehension Dataset for Curriculum Learning. In *Proceedings of Machine Learning Research* (Vol. 101, pp. 742–757). PMLR. https://proceedings.mlr.press/v101/liang19a.html

Liu, H., Liu, J., Cui, L., Teng, Z., Duan, N., Zhou, M., & Zhang, Y. (2023). LogiQA 2.0 - An Improved Dataset for Logical Reasoning in Natural Language Understanding. *IEEE/ACM Transactions on Audio Speech and Language Processing*, *31*, 2947–2962. https://doi.org/10.1109/TASLP.2023.3293046

Liu, J., Cui, L., Liu, H., Huang, D., Wang, Y., & Zhang, Y. (2020). LogiQA: A Challenge Dataset for Machine Reading Comprehension with Logical Reasoning. *IJCAI International*

Joint Conference on Artificial Intelligence, 2021-January, 3622–3628. https://doi.org/10.24963/ijcai.2020/501

Liu, P., Yuan, W., Fu, J., Jiang, Z., Hayashi, H., & Neubig, G. (2023). Pre-train, Prompt, and Predict: A Systematic Survey of Prompting Methods in Natural Language Processing. *ACM Computing Surveys*, *55*(9). https://doi.org/10.1145/3560815

Liu, S., Zhang, X., Zhang, S., Wang, H., & Zhang, W. (2019). Neural Machine Reading Comprehension: Methods and Trends. *Applied Sciences 2019, Vol. 9, Page 3698*, *9*(18), 3698. https://doi.org/10.3390/APP9183698

Liu, Y., Ott, M., Goyal, N., Du, J., Joshi, M., Chen, D., Levy, O., Lewis, M., Zettlemoyer, L., & Stoyanov, V. (2019). RoBERTa: A Robustly Optimized BERT Pretraining Approach. *ArXiv:1907.11692*. http://arxiv.org/abs/1907.11692

Ma, F., Zhang, C., Ren, L., Wang, J., Wang, Q., Wu, W., Quan, X., & Song, D. (2022). XPrompt: Exploring the Extreme of Prompt Tuning. *Proceedings of the 2022 Conference on Empirical Methods in Natural Language Processing, EMNLP 2022*, 11033–11047. https://doi.org/10.18653/v1/2022.emnlp-main.758

Meng, X., Song, Y., Bai, Q., & Wang, T. (2023). CBKI: A confidence-based knowledge integration framework for multi-choice machine reading comprehension. *Knowledge-Based Systems*, 110796. https://doi.org/10.1016/J.KNOSYS.2023.110796

Min, S., Lyu, X., Holtzman, A., Artetxe, M., Lewis, M., Hajishirzi, H., & Zettlemoyer, L. (2022). Rethinking the Role of Demonstrations: What Makes In-Context Learning Work? *Proceedings of the 2022 Conference on Empirical Methods in Natural Language Processing, EMNLP 2022*, 11048–11064. https://doi.org/10.18653/v1/2022.emnlp-main.759

Mohammadi, A., Ramezani, R., & Baraani, A. (2023). Topic-aware multi-hop machine reading comprehension using weighted graphs. *Expert Systems with Applications*, *224*, 119873. https://doi.org/10.1016/J.ESWA.2023.119873

Onishi, T., Wang, H., Bansal, M., Gimpel, K., & McAllester, D. (2016). Who did What: A Large-Scale Person-Centered Cloze Dataset. *EMNLP 2016 - Conference on Empirical Methods in Natural Language Processing, Proceedings*, 2230–2235. https://doi.org/10.18653/v1/d16-1241

OpenAI, :, Achiam, J., Adler, S., Agarwal, S., Ahmad, L., Akkaya, I., Aleman, F. L., Almeida, D., Altenschmidt, J., Altman, S., Anadkat, S., Avila, R., Babuschkin, I., Balaji, S., Balcom, V., Baltescu, P., Bao, H., Bavarian, M., … Zoph, B. (2023). *GPT-4 Technical Report*. https://arxiv.org/abs/2303.08774v4

Ostermann, S., Modi, A., Roth, M., Thater, S., & Pinkal, M. (2018). *MCScript: A Novel Dataset for Assessing Machine Comprehension Using Script Knowledge*. https://aclanthology.org/L18-1564

Ostermann, S., Roth, M., & Pinkal, M. (2019). MCScript2.0: A Machine Comprehension Corpus Focused on Script Events and Participants. *SEM@NAACL-HLT 2019 - 8th Joint Conference on Lexical and Computational Semantics*, 103–117. https://doi.org/10.18653/V1/S19-1012

Ouyang, L., Wu, J., Jiang, X., Almeida, D., Wainwright, C. L., Mishkin, P., Zhang, C., Agarwal, S., Slama, K., Ray, A., Schulman, J., Hilton, J., Kelton, F., Miller, L., Simens, M., Askell, A., Welinder, P., Christiano, P., Leike, J., & Lowe, R. (2022). Training language models to follow instructions with human feedback. *Advances in Neural Information Processing Systems*, *35*. https://arxiv.org/abs/2203.02155v1

Pan, X., Sun, K., Yu, D., Chen, J., Ji, H., Cardie, C., & Yu, D. (2019). Improving Question Answering with External Knowledge. *Proceedings of the 2nd Workshop on Machine Reading for Question Answering*, 27–37. https://doi.org/10.18653/V1/D19-5804

Pappas, D., Androutsopoulos, I., & Papageorgiou, H. (2018). BioRead: A New Dataset for Biomedical Reading Comprehension. *Proceedings of the Eleventh International Conference on Language Resources and Evaluation (LREC-2018)*.

Paulus, R., Xiong, C., & Socher, R. (2017). A Deep Reinforced Model for Abstractive Summarization. *ArXiv:1705.04304*. http://arxiv.org/abs/1705.04304

Peng, W., Hu, Y., Yu, J., Xing, L., & Xie, Y. (2021). APER: AdaPtive Evidence-driven Reasoning Network for machine reading comprehension with unanswerable questions. *Knowledge-Based Systems*, *229*, 107364. https://doi.org/10.1016/J.KNOSYS.2021.107364

Peters, M. E., Neumann, M., Logan, R. L., Schwartz, R., Joshi, V., Singh, S., & Smith, N. A. (2019). Knowledge Enhanced Contextual Word Representations. *EMNLP-IJCNLP 2019 - 2019 Conference on Empirical Methods in Natural Language Processing and 9th International Joint Conference on Natural Language Processing, Proceedings of the Conference*, 43–54. https://doi.org/10.18653/V1/D19-1005

Qiu, B., Chen, X., Xu, J., & Sun, Y. (2019). *A Survey on Neural Machine Reading Comprehension*. https://arxiv.org/abs/1906.03824v1

Raffel, C., Shazeer, N., Roberts, A., Lee, K., Narang, S., Matena, M., Zhou, Y., Li, W., & Liu, P. J. (2019). Exploring the Limits of Transfer Learning with a Unified Text-to-Text Transformer. *Journal of Machine Learning Research*, *21*, 1–67. https://arxiv.org/abs/1910.10683v3

Rogers, A., Kovaleva, O., Downey, M., & Rumshisky, A. (2020). Getting Closer to AI Complete Question Answering: A Set of Prerequisite Real Tasks. *Proceedings of the AAAI Conference on Artificial Intelligence*, *34*(05), 8722–8731. https://doi.org/10.1609/AAAI.V34I05.6398

Stavropoulos, P., Pappas, D., Androutsopoulos, I., & McDonald, R. (2020). BioMRC: A Dataset for Biomedical Machine Reading Comprehension. *Proceedings of the Annual Meeting of the Association for Computational Linguistics*, 140–149. https://doi.org/10.18653/V1/2020.BIONLP-1.15

Sun, K., Yu, D., Chen, J., Yu, D., Choi, Y., & Cardie, C. (2019). DREAM: A Challenge Data Set and Models for Dialogue-Based Reading Comprehension. *Transactions of the Association for Computational Linguistics*, *7*, 217–231. https://doi.org/10.1162/tacl_a_00264

Šuster, S., & Daelemans, W. (2018). CliCR: A Dataset of Clinical Case Reports for Machine Reading Comprehension. *ArXiv:1803.09720*. http://arxiv.org/abs/1803.09720

Tang, T., Li, J., Zhao, W. X., & Wen, J. R. (2022). Context-Tuning: Learning Contextualized Prompts for Natural Language Generation. *Proceedings - International Conference on Computational Linguistics, COLING*, *29*(1), 6340–6354. https://arxiv.org/abs/2201.08670v2

Tapaswi, M., Zhu, Y., Stiefelhagen, R., Torralba, A., Urtasun, R., & Fidler, S. (2015). MovieQA: Understanding Stories in Movies through Question-Answering. *Proceedings of the IEEE Computer Society Conference on Computer Vision and Pattern Recognition*, *2016-December*, 4631–4640. https://doi.org/10.1109/CVPR.2016.501

Team, G., Anil, R., Borgeaud, S., Wu, Y., Alayrac, J.-B., Yu, J., Soricut, R., Schalkwyk, J., Dai, A. M., Hauth, A., Millican, K., Silver, D., Petrov, S., Johnson, M., Antonoglou, I., Schrittwieser, J., Glaese, A., Chen, J., Pitler, E., … Vinyals, O. (2023). *Gemini: A Family of Highly Capable Multimodal Models*. https://arxiv.org/abs/2312.11805v1

Vaswani, A., Shazeer, N., Parmar, N., Uszkoreit, J., Jones, L., Gomez, A. N., Kaiser, Ł., & Polosukhin, I. (2017). Attention is All you Need. *Advances in Neural Information Processing Systems 30*, 5998–6008. http://papers.nips.cc/paper/7181-attention-is-all-you-need

Wang, A., Pruksachatkun, Y., Nangia, N., Singh, A., Michael, J., Hill, F., Levy, O., & Bowman, S. R. (2019). SuperGLUE: A Stickier Benchmark for General-Purpose Language

Understanding Systems. *Advances in Neural Information Processing Systems*, *32*. https://arxiv.org/abs/1905.00537v3

Wang, S., Sun, Y., Xiang, Y., Wu, Z., Ding, S., Gong, W., Feng, S., Shang, J., Zhao, Y., Pang, C., Liu, J., Chen, X., Lu, Y., Liu, W., Wang, X., Bai, Y., Chen, Q., Zhao, L., Li, S., … Wang, H. (2021). *ERNIE 3.0 Titan: Exploring Larger-scale Knowledge Enhanced Pre-training for Language Understanding and Generation*. https://arxiv.org/abs/2107.02137v1

Wei, J., Bosma, M., Zhao, V. Y., Guu, K., Yu, A. W., Lester, B., Du, N., Dai, A. M., & Le, Q. V. (2021). Finetuned Language Models Are Zero-Shot Learners. *ICLR 2022 - 10th International Conference on Learning Representations*. https://arxiv.org/abs/2109.01652v5

Wei, J., Wang, X., Schuurmans, D., Bosma, M., Ichter, B., Xia, F., Chi, E. H., Le, Q. V., & Zhou, D. (2022). Chain-of-Thought Prompting Elicits Reasoning in Large Language Models. *Advances in Neural Information Processing Systems*, *35*. https://arxiv.org/abs/2201.11903v6

Welbl, J., Liu, N. F., & Gardner, M. (2017). Crowdsourcing Multiple Choice Science Questions. *3rd Workshop on Noisy User-Generated Text, W-NUT 2017 - Proceedings of the Workshop*, 94–106. https://doi.org/10.18653/V1/W17-4413

Welbl, J., Stenetorp, P., & Riedel, S. (2018). Constructing Datasets for Multi-hop Reading Comprehension Across Documents. *Transactions of the Association for Computational Linguistics*, *6*, 287–302. https://doi.org/10.1162/TACL_A_00021

Xie, Q., Lai, G., Dai, Z., & Hovy, E. (2018). Large-scale Cloze Test Dataset Created by Teachers. *Proceedings of the 2018 Conference on Empirical Methods in Natural Language Processing, EMNLP 2018*, 2344–2356. https://doi.org/10.18653/V1/D18-1257

Yagcioglu, S., Erdem, A., Erdem, E., & Ikizler-Cinbis, N. (2018). RecipeQA: A Challenge Dataset for Multimodal Comprehension of Cooking Recipes. *Proceedings of the 2018 Conference on Empirical Methods in Natural Language Processing, EMNLP 2018*, 1358–1368. https://doi.org/10.18653/V1/D18-1166

Yamada, I., Asai, A., Shindo, H., Takeda, H., & Matsumoto, Y. (2020). LUKE: Deep Contextualized Entity Representations with Entity-aware Self-attention. *EMNLP 2020 - 2020 Conference on Empirical Methods in Natural Language Processing, Proceedings of the Conference*, 6442–6454. https://doi.org/10.18653/V1/2020.EMNLP-MAIN.523

Yang, Z., Dai, Z., Yang, Y., Carbonell, J., Salakhutdinov, R. R., & Le, Q. V. (2019). Xlnet: Generalized autoregressive pretraining for language understanding. *Advances in Neural Information Processing Systems*, *32*.

Yu, W., Jiang, Z., Dong, Y., & Feng, J. (2020). ReClor: A Reading Comprehension Dataset Requiring Logical Reasoning. *8th International Conference on Learning Representations, ICLR 2020*. https://arxiv.org/abs/2002.04326v3

Zaheer, M., Guruganesh, G., Dubey, A., Ainslie, J., Alberti, C., Ontanon, S., Pham, P., Ravula, A., Wang, Q., Yang, L., & Research, A. G. (2020). Big Bird: Transformers for Longer Sequences. *Advances in Neural Information Processing Systems*, *33*, 17283–17297.

Zeng, C., Li, S., Li, Q., Hu, J., & Hu, J. (2020). A Survey on Machine Reading Comprehension: Tasks, Evaluation Metrics and Benchmark Datasets. *Applied Sciences (Switzerland)*, *10*(21), 1–57. https://doi.org/10.3390/app10217640

Zhang, S., Liu, X., Liu, J., Gao, J., Duh, K., & Van Durme, B. (2018). ReCoRD: Bridging the Gap between Human and Machine Commonsense Reading Comprehension. *ArXiv:1810.12885*. http://arxiv.org/abs/1810.12885

Zhang, Z., Han, X., Liu, Z., Jiang, X., Sun, M., & Liu, Q. (2019). ERNIE: Enhanced Language Representation with Informative Entities. *ACL 2019 - 57th Annual Meeting of the Association for Computational Linguistics, Proceedings of the Conference*, 1441–1451.